\pdfoutput=1

\documentclass[11pt]{article}

\usepackage[final]{acl}

\usepackage{times}
\usepackage{latexsym}

\usepackage[T1]{fontenc}

\usepackage[utf8]{inputenc}

\usepackage{microtype}

\usepackage{inconsolata}

\usepackage{graphicx}

\usepackage{multirow}
\usepackage{amssymb}
\usepackage{makecell}
\usepackage{amsmath}
\usepackage{relsize}
\usepackage{booktabs}
\usepackage{bm}
\usepackage{hyperref}

\usepackage{xcolor}
\usepackage{wrapfig} 
\usepackage{colortbl}
\usepackage[most]{tcolorbox}

\definecolor{wkyellow}{RGB}{255,241,177}
\definecolor{lightgray}{HTML}{CCE5FF}
\definecolor{lightgray}{gray}{0.9}
\definecolor{goodblue}{HTML}{0071bc}

\makeatletter
\renewcommand{\maketag@@@}[1]{\hbox{\m@th\normalsize\normalfont#1}}%
\makeatother

\title{Single-to-mix Modality Alignment with Multimodal Large Language Model for Document Image Machine Translation}

\author{
\normalsize{Yupu Liang}\textsuperscript{1,2},
\normalsize{Yaping Zhang}\textsuperscript{1,2} \thanks{\ \ Corresponding author.}\ ,
\normalsize{Zhiyang Zhang}\textsuperscript{1,2},\\
{\bf \normalsize{Yang Zhao}\textsuperscript{1,2}},
{\bf \normalsize{Lu Xiang}\textsuperscript{1,2}},
{\bf \normalsize{Chengqing Zong}\textsuperscript{1,2}},
{\bf \normalsize{Yu Zhou}\textsuperscript{1,3}}\\
\textsuperscript{1} \normalsize{State Key Laboratory of Multimodal Artificial Intelligence Systems (MAIS),} \\ \normalsize{Institute of Automation, Chinese Academy of Sciences, Beijing, China} \\
\textsuperscript{2} \normalsize{School of Artificial Intelligence, University of Chinese Academy of Sciences, Beijing, China} \\
\textsuperscript{3} \normalsize{Fanyu AI Laboratory, Zhongke Fanyu Technology Co., Ltd, Beijing, China} \\
\text{\small{\{liangyupu2021, zhangzhiyang2020\}@ia.ac.cn, }}\small{\{yaping.zhang, yang.zhao, lu.xiang, cqzong, yzhou\}@nlpr.ia.ac.cn}
}

\begin{document}
\maketitle
\begin{abstract}

Document Image Machine Translation (DIMT) aims to translate text within document images, facing generalization challenges due to limited training data and the complex interplay between visual and textual information.
To address these challenges, we introduce M4Doc, a novel single-to-mix modality alignment framework leveraging Multimodal Large Language Models (MLLMs). 
M4Doc aligns an image-only encoder with the multimodal representations of an MLLM, pre-trained on large-scale document image datasets.
This alignment enables a lightweight DIMT model to learn crucial visual-textual correlations during training.
During inference, M4Doc bypasses the MLLM, maintaining computational efficiency while benefiting from its multimodal knowledge.
Comprehensive experiments demonstrate substantial improvements in translation quality, especially in cross-domain generalization and challenging document image scenarios.\footnote{Our code is available at: \url{https://github.com/liangyupu/M4Doc}}

\end{abstract}

\section{Introduction}   

\begin{figure}[t]
    \centering
    \includegraphics[width=\columnwidth]{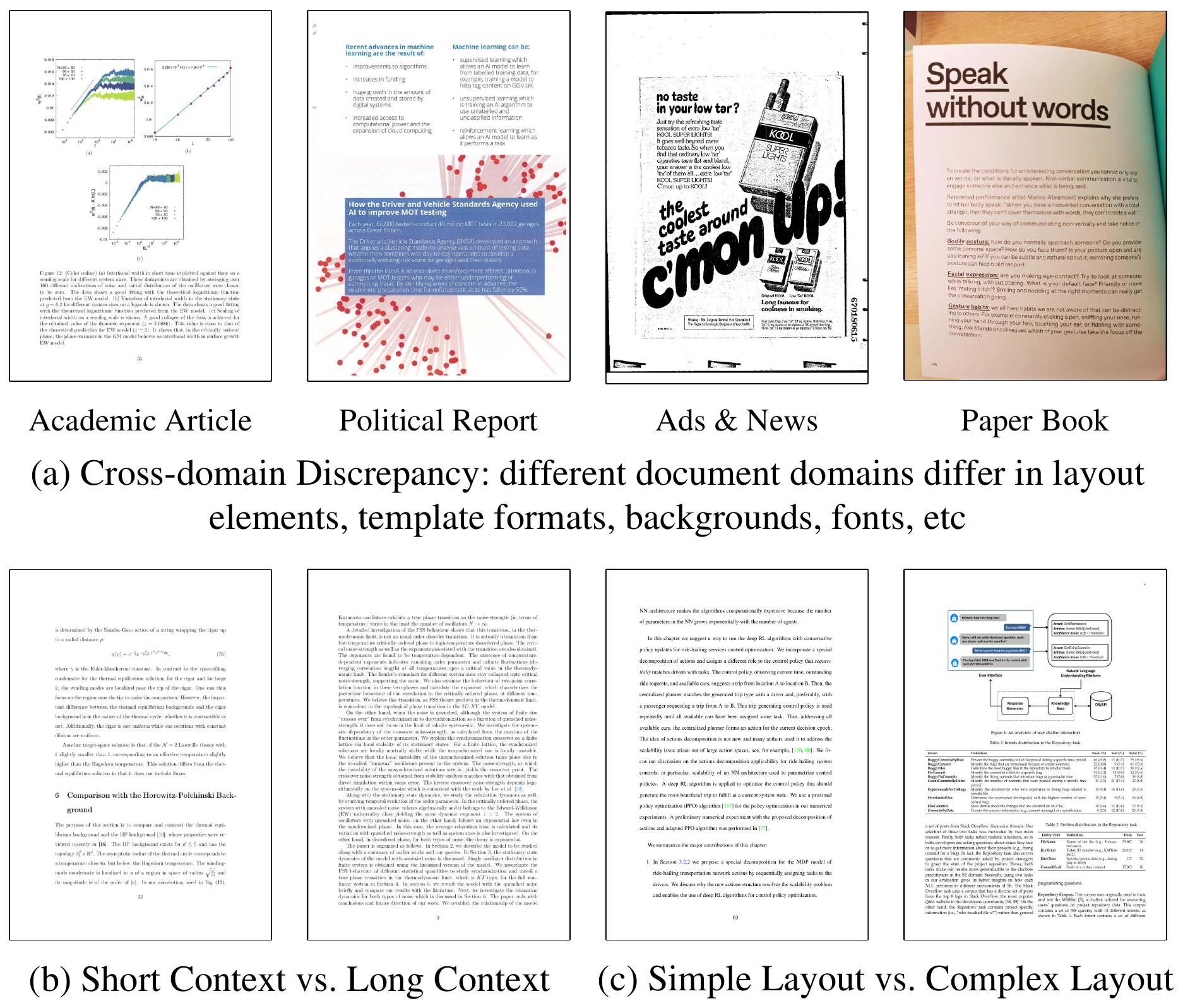}
    \caption{Different test scenarios of end-to-end DIMT.}
    \label{figure: introduction}
\end{figure}

Document Image Machine Translation (DIMT) aims to translate text within document images from one language to another while preserving the logical layout \citep{liang-etal-2024-document}.
With vast amounts of information stored in document images (e.g., academic papers, magazines, scanned documents, Figure~\ref{figure: introduction}), DIMT has gained increasing attention as a critical sub-task of visual document understanding in the era of multimodal large language models\citep{ye2023mplug, zhang2023llavar, hu2024mplug, yu2024texthawk}.

Recent advancements in DIMT can be categorized into two primary approaches: (1) Cascade systems \citep{hinami2021towards, sable2023doc, yao2023docxchain, zhang2023novel}, which employ multiple models sequentially and encounter issues such as structural redundancy, error propagation, and high latency.
(2) End-to-end methods \citep{jain2021image, ma2022improving, zhu-etal-2023-peit, zhang-etal-2023-layoutdit, liang-etal-2024-document, zhang-etal-2025-chaotic, DBLP:journals/pami/ZhangZLMXZZZ25}, which streamline the process by optimizing a unified training objective, thus enhancing structural efficiency.

\begin{table}[t]
\footnotesize
\centering
\begin{tabular}{ll|cc}
\toprule
\multicolumn{2}{c|}{}                      & \textbf{BLEU}  & \textbf{BLEU-PT} \\ \midrule
\multicolumn{1}{l|}{1} & DoTA Test Set & 38.68 & 42.34   \\
\multicolumn{1}{l|}{2} & Cross-domain      & 12.64 & 15.03   \\
\multicolumn{1}{l|}{3} & Long Context      & 34.85 & 33.73   \\
\multicolumn{1}{l|}{4} & Complex Layout    & 30.30 & 35.16   \\ \bottomrule
\end{tabular}
\caption{Scores of the end-to-end DIMT model on different test scenarios. The model is trained on the DoTA dataset. \textit{Cross-domain}, \textit{Long Context}, and \textit{Complex Layout} mean testing on the DITrans political report subset, the DoTA long context subset, and the complex layout subset, separately.}
\label{table: introduction}
\end{table}

However, both cascade and end-to-end approaches are hindered by the lack of large and diverse DIMT datasets, which limits their ability to generalize to new types of documents.
This limited generalization is evident in performance drops across several key scenarios, as shown in Table~\ref{table: introduction}:
(1) \textbf{cross-domain generalization}: the model, trained on the DoTA dataset \citep{liang-etal-2024-document}, achieves a BLEU score of 38.68 on the original test set, but only 12.64 on the DITrans Political Report test set \citep{zhang-etal-2023-layoutdit} in a cross-domain zero-shot scenario, which includes document images with varying layouts, fonts, and background.
(2) \textbf{long context generalization}: there is a decrease of 3.83 BLEU score when testing on the same dataset but with a long context subset, containing document images with more than 750 English words.
(3) \textbf{complex layout generalization}: the model's performance drops by 8.38 BLEU when the test set includes more images with complex layouts.\footnote{The criteria for complex layouts is in Appendix~\ref{sec: appendix dataset setting}.}

Recent Multimodal Large Language Models (MLLMs), pre-trained on extensive datasets of images and text, have demonstrated impressive generalization across various domains, contexts, and layouts \citep{ye2023mplug, zhang2023llavar, hu2024mplug, yu2024texthawk}. 
While MLLMs hold great promise for DIMT \citep{liu2024llavanext, liu2024textmonkey}, their large size and computational demands make them difficult to use directly, especially in resource-constrained environments.\footnote{More details on fine-tuning MLLMs for DIMT are in Appendix~\ref{sec: appendix fine-tune mllm}.}

To address these limitations, we propose \textbf{M4Doc} (single-to-mix \textbf{M}odality alignment with \textbf{M}ultimodal large language \textbf{M}odel for \textbf{Doc}ument image \textbf{M}achine translation), a novel framework that leverages the strong generalization capabilities of MLLMs to enhance the performance and efficiency of smaller DIMT models through a single-to-mix modality alignment strategy.
This strategy aligns an image-only representation with the MLLM's rich multimodal representations, effectively transferring knowledge from the MLLM.
Specifically, a novel single-to-mix modality alignment encoder is designed as a bridge to connect the MLLM and the DIMT model.
This encoder with \textbf{only image} input learns to align with the mix-modality representation of MLLM, using both \textbf{image and text} as inputs.
The alignment encoder can serve as an alternative to the MLLM and provide mix-modality information to the DIMT model in the inference stage.
A major advantage of this approach is that it requires aligning the DIMT model with the MLLM only during training, allowing the use of a smaller model during inference, which achieves the trade-off between performance and inference speed.

Our contributions are summarized as follows:

\begin{itemize}
\item A novel method, M4Doc, has been proposed, which uses the pre-trained knowledge of the MLLM to assist a small DIMT model in the training stage and achieves the trade-off between translation quality and inference speed.
\item A new approach, single-to-mix alignment, has been developed, which only takes images as input and aligns with the mix-modality representation of the MLLM. 
\item Extensive experiments demonstrate the effectiveness of the proposed method, and the DIMT model's performances on cross-domain, long context, and complex layout scenarios are also improved.
\end{itemize}

\begin{figure*}[t]
    \centering
    \includegraphics[width=2\columnwidth]{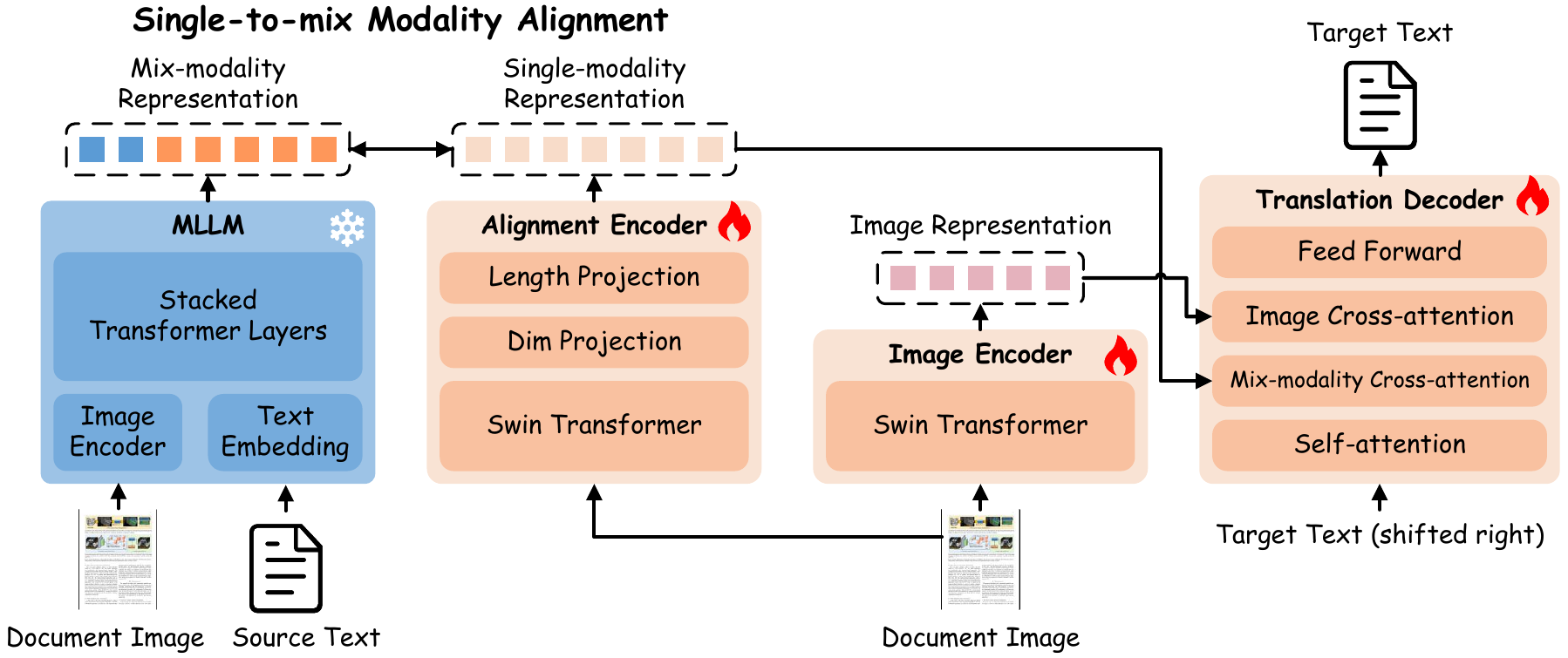}
    \caption{The diagram of the proposed M4Doc. During training, the alignment encoder learns to align with the MLLM's mix-modality representation with single-modality input. The MLLM is frozen, while the other modules remain trainable. During inference, the MLLM is discarded for faster inference speed, while the alignment encoder provides aligned mix-modality information to guide the translation decoder.}
    \label{figure: model}
\end{figure*}

\section{M4Doc Mehod}
In this section, we introduce M4Doc, a novel single-to-mix modality alignment framework designed to enhance DIMT by leveraging the MLLMs.
The model architecture of M4Doc is illustrated in Figure~\ref{figure: model}.
The key idea of M4Doc is to align the representations of an image-only encoder with the rich multimodal representations of an MLLM during training, enabling a lightweight DIMT model to effectively capture the interplay between textual and visual features.
The whole model contains an MLLM, an alignment encoder, an image encoder, and a translation decoder.
In the training stage, the alignment encoder simultaneously learns to align with the MLLM and provides mix-modality information to the translation decoder.
In the inference stage, the alignment encoder serves as an alternative to the MLLM and continues providing mix-modality information with only image input.

\subsection{Mix-modality Representation Extraction}
The MLLM acts as a guide for the alignment encoder to provide mix-modality information with image and text inputs.
We input the image $\bm{I} \in \mathbb{R}^{\mathrm{H} \times \mathrm{W} \times 3}$ and corresponding ground truth source language text $\bm{X} = \{x_1, x_2, ..., x_m\}$ into the MLLM.
The input format is \texttt{<System Prompt> <Image Token> <User Prompt> <Source Text>}, which is the same as the format used in the MLLM pre-training.
\texttt{<System Prompt>} and \texttt{<User Prompt>} are also the same as those used by the MLLM in the OCR task.
\texttt{<Source Text>} is the ground truth OCR text of the corresponding image.
We can get the mix-modality representation\footnote{The last layer's output hidden states of the MLLM.} $\bm{H}_{\mathrm{MLLM}} \in \mathbb{R}^{l_{\mathrm{MLLM}} \times d_{\mathrm{MLLM}}}$, which can be formulated as:
\begin{equation}
\small
    \bm{H}_{\mathrm{MLLM}} = \mathrm{MLLM} (\bm{I}, \bm{X})
\end{equation}
where $l_{\mathrm{MLLM}}$ and $d_{\mathrm{MLLM}}$ are the sequence length and dimension of the MLLM.

\subsection{Single-to-mix Modality Alignment}
The alignment encoder bridges the gap between the single-modality (image-only) input and the mix-modality (image and text) representations of the MLLM. 
Using a pre-trained Swin Transformer \citep{blecher2024nougat}, the alignment encoder extracts visual features $\bm{H}_{\mathrm{Swin}} \in \mathbb{R}^{l_{\mathrm{Swin}} \times d_{\mathrm{Swin}}}$  from the image $\bm{I}$:
\begin{equation}
\small
    \bm{H}_{\mathrm{Swin}} = \mathrm{Swin} (\bm{I})
\end{equation}
To match the dimensions of the MLLM output, two Feed Forward Networks (FFNs) are used to project $\bm{H}_{\mathrm{Swin}}$ to $\bm{H}_{\mathrm{Align}} \in \mathbb{R}^{l_{\mathrm{MLLM}} \times d_{\mathrm{MLLM}}}$:
\begin{equation}
\small
    \bm{H}_{\mathrm{Align}} = \mathrm{FFN}_{\mathrm{length}} (\mathrm{FFN}_{\mathrm{dim}} (\bm{H}_{\mathrm{Swin}})^T)^T
\end{equation}
After this process, an alignment loss guides the alignment encoder to mimic the mix-modality representation $\bm{H}_{\mathrm{Align}} \in \mathbb{R}^{l_{\mathrm{MLLM}} \times d_{\mathrm{MLLM}}}$\footnote{The effect of different alignment loss functions can be found in Appendix~\ref{sec: loss function}.}:
\begin{equation}
\small
    \mathcal{L}_{\mathrm{align}} = 1 - \mathrm{Cos} (\bm{H}_{\mathrm{MLLM}}, \bm{H}_{\mathrm{Align}})
\end{equation}
where $\mathrm{Cos}$ is the cosine similarity of two tensors.

\subsection{Aligned Mix-modality Guided Translation}
The image encoder encodes the input image $\bm{I}$ to its semantic representation $\bm{H}_{\mathrm{Image}} \in \mathbb{R}^{l_{\mathrm{Image}} \times d_{\mathrm{Image}}}$.
We also use a pre-trained Swin Transformer \citep{blecher2024nougat} to construct the image encoder.
$\bm{H}_{\mathrm{Image}}$ is calculated as follows:
\begin{equation}
\small
    \bm{H}_{\mathrm{Image}} = \mathrm{Encoder}_{\mathrm{Image}} (\bm{I})
\end{equation}
where $l_{\mathrm{Image}}$ is the number of output vectors and $d_{\mathrm{Image}}$ is the vectors' dimension.

The translation decoder is aimed to generate target language text under the guidance of the alignment encoder and image encoder.
We modify the vanilla Transformer's decoder \citep{vaswani2017attention} by incorporating a mix-modality cross-attention module and an image cross-attention module in each layer to receive representations from the alignment encoder and the image encoder.
At each decoding timestep $t$, the translation decoder takes $\bm{H}_{\mathrm{Align}}$, $\bm{H}_{\mathrm{Image}}$ and generated target tokens $y_{<t} = \{y_1, y_2, ..., y_{t-1}\}$ as input and outputs the probability distribution of next target token $y_t$.
This process can be defined as:
\begin{equation}
\small
    p(y_t | y_{<t}, \bm{I}, \bm{X}) = \mathrm{Decoder} (y_{<t}, \bm{H}_{\mathrm{Align}}, \bm{H}_{\mathrm{Image}})
\end{equation}
where $\bm{H}_{\mathrm{Align}}$ and $\bm{H}_{\mathrm{Image}}$ both need to be converted to the same dimension of the decoder through two FFNs which are not shown in Figure~\ref{figure: model} for simplicity.

The translation loss is as follows:
\begin{equation}
\small
    \mathcal{L}_{\mathrm{trans}} = -\sum_{t=1}^{n} \log p(y_t | y_{<t}, \bm{I}, \bm{X}; \bm{\theta})
\end{equation}
where $\bm{\theta}$ is the parameters of the alignment encoder, image encoder, and translation decoder.

\subsection{Training \& Inference Strategy}
In the training stage, the Swin Transformer modules of the alignment encoder and image encoder are initialized from the pre-trained OCR model's encoder.
The translation decoder's FFN, image cross-attention, and self-attention modules are initialized from the pre-trained text translation model's decoder.
The other parts are randomly initialized.
The parameters of the MLLM are frozen.

The total loss of M4Doc is as follows:
\begin{equation}
    \small
    \mathcal{L} = \alpha \times \mathcal{L}_{\mathrm{align}} + \mathcal{L}_{\mathrm{trans}}
\end{equation}
where $\alpha$ is a hyperparameter.\footnote{The effect of different hyperparameters can be found in Appendix~\ref{sec: hyperparameter}.}

In the inference stage, as shown in Figure~\ref{figure: model}, only the alignment encoder, image encoder, and translation decoder are involved, which contain much fewer parameters compared with the MLLM.
Furthermore, due to the introduction of the alignment encoder aligning with the MLLM during the training stage, the entire model maintains high translation quality while achieving fast inference speed.

\section{Experiments}
\subsection{Dataset \& Metrics}
Our models are comprehensively evaluated on two public benchmarks DoTA \citep{liang-etal-2024-document} and DITrans \citep{zhang2023novel}, under academic article and political report scenarios.
Detailed dataset setting can be seen in Appendix~\ref{sec: appendix dataset setting}.

We thoroughly evaluate the models' capabilities in three aspects: (1) \textbf{full-text translation}, which means the translation quality of all the text in the image - BLEU and COMET \citep{rei-etal-2020-comet}. (2) \textbf{plain-text translation}, which means the translation quality of the text after removing formulas and tables - BLEU-PT. (3) \textbf{structure preserving}, which means the model's ability to restore the layout structure of the document images - STEDS (Structure Tree-Edit-Distance-based Similarity).

We calculate BLEU, BLEU-PT, and STEDS the same as \citet{liang-etal-2024-document}.
For COMET calculation, due to the original COMET's inability to process long texts, we first used Trankit \citep{nguyen-etal-2021-trankit} to segment the source and translated texts into sentences, then used Sentalign \citep{steingrimsson-etal-2023-sentalign} for sentence-level alignment, and finally calculated the average of COMET score in reference-free mode.\footnote{The COMET model we used is \href{https://huggingface.co/Unbabel/wmt22-comet-da}{wmt22-comet-da}.}


\begin{table*}[t]
\centering
\footnotesize
\resizebox{\linewidth}{!}{
\begin{tabular}{llcccccccccc}
\toprule
\multicolumn{2}{c|}{\multirow{2}{*}{}}                            & \multicolumn{4}{c|}{\textbf{DoTA}}                                                     & \multicolumn{4}{c|}{\textbf{DITrans}}                                                  & \multirow{2}{*}{\textbf{\begin{tabular}[c]{@{}c@{}}\# Params\\ (M)\end{tabular}}} & \multirow{2}{*}{\textbf{\begin{tabular}[c]{@{}c@{}}Time\\ (s/page)\end{tabular}}} \\
\multicolumn{2}{c|}{}                                             & \textbf{B}     & \textbf{C}     & \textbf{BP}    & \multicolumn{1}{c|}{\textbf{S}}     & \textbf{B}     & \textbf{C}     & \textbf{BP}    & \multicolumn{1}{c|}{\textbf{S}}     &                                                                                  &                                                                                   \\ \midrule
\multicolumn{1}{l|}{1}  & \multicolumn{1}{l|}{Text-only MT}       & 47.61          & 67.51          & 54.16          & \multicolumn{1}{c|}{92.89}          & 21.50          & 48.76          & 22.55          & \multicolumn{1}{c|}{86.96}          & 99.5                                                                             & 8.81                                                                              \\ \midrule
\multicolumn{12}{c}{\textbf{Cascade Baselines}}                                                                                                                                                                                                                                                                                                                                                                            \\ \midrule
\multicolumn{1}{l|}{2}  & \multicolumn{1}{l|}{LARDIT}                & 35.58          & 54.48          & 41.75          & \multicolumn{1}{c|}{75.83}          & 14.66          & 30.16          & 16.58          & \multicolumn{1}{c|}{57.77}          & 99.5 + $\theta_1$                                                                           & 12.46                                                                             \\
\multicolumn{1}{l|}{3}  & \multicolumn{1}{l|}{Nougat-trans}                 & 43.37          & 65.25          & 50.79          & \multicolumn{1}{c|}{88.16}          & 18.39          & 35.80          & 19.21          & \multicolumn{1}{c|}{52.12}          & 346.9                                                                            & 17.03                                                                             \\ \midrule
\multicolumn{12}{c}{\textbf{End-to-end TIMT Baselines (Document-level)}}                                                                                                                                                                                                                                                                                                                                                \\ \midrule
\multicolumn{1}{l|}{4}  & \multicolumn{1}{l|}{ItNet}              & 3.84           & 21.94          & 2.27           & \multicolumn{1}{c|}{48.46}          & 1.64           & 21.52          & 1.71           & \multicolumn{1}{c|}{41.63}          & 97.5                                                                             & 8.43                                                                              \\
\multicolumn{1}{l|}{5}  & \multicolumn{1}{l|}{E2ETIT}             & 1.51           & 20.80          & 1.69           & \multicolumn{1}{c|}{32.90}          & 2.71           & 23.45          & 2.83           & \multicolumn{1}{c|}{40.53}          & 122.0                                                                            & 8.19                                                                              \\
\multicolumn{1}{l|}{6}  & \multicolumn{1}{l|}{PEIT}               & 5.81           & 24.98          & 4.52           & \multicolumn{1}{c|}{55.79}          & 4.13           & 21.98          & 4.21           & \multicolumn{1}{c|}{41.59}          & 135.1                                                                            & 2.57                                                                              \\ \midrule
\multicolumn{12}{c}{\textbf{End-to-end TIMT Baselines (Text-line-level)}}                                                                                                                                                                                                                                                                                                                                                      \\ \midrule
\multicolumn{1}{l|}{7}  & \multicolumn{1}{l|}{ItNet}              & 21.75          & 43.29          & 23.52          & \multicolumn{1}{c|}{75.83}          & 6.16           & 28.82          & 8.77           & \multicolumn{1}{c|}{57.77}          & 97.5 + $\theta_2$                                                                           & 7.20                                                                              \\
\multicolumn{1}{l|}{8}  & \multicolumn{1}{l|}{E2ETIT}             & 17.42          & 38.25          & 17.74          & \multicolumn{1}{c|}{75.83}          & 6.72           & 28.55          & 7.81           & \multicolumn{1}{c|}{57.77}          & 122.0 + $\theta_2$                                                                          & 7.59                                                                              \\
\multicolumn{1}{l|}{9}  & \multicolumn{1}{l|}{PEIT}               & 27.43          & 44.08          & 31.29          & \multicolumn{1}{c|}{75.83}          & 9.08           & 26.18          & 9.38           & \multicolumn{1}{c|}{57.77}          & 135.1 + $\theta_2$                                                                          & 2.42                                                                              \\ \midrule
\multicolumn{12}{c}{\textbf{Knowledge   Distillation Baselines}}                                                                                                                                                                                                                                                                                                                                                           \\ \midrule
\multicolumn{1}{l|}{10} & \multicolumn{1}{l|}{Seq-KD}             & 34.42          & 53.54          & 36.63          & \multicolumn{1}{c|}{82.51}          & 10.58          & 25.89          & 11.38          & \multicolumn{1}{c|}{56.92}          & 212.4                                                                            & 9.76                                                                              \\
\multicolumn{1}{l|}{11} & \multicolumn{1}{l|}{MTKD}               & 37.32          & 60.32          & 39.96          & \multicolumn{1}{c|}{82.28}          & 13.24          & 29.33          & 15.33          & \multicolumn{1}{c|}{59.58}          & 212.4                                                                            & 9.56                                                                              \\
\multicolumn{1}{l|}{12} & \multicolumn{1}{l|}{RD (Original)}      & 5.13           & 23.86          & 3.85           & \multicolumn{1}{c|}{53.06}          & 0.53           & 24.37          & 0.56           & \multicolumn{1}{c|}{40.07}          & 212.4                                                                            & 8.38                                                                              \\
\multicolumn{1}{l|}{13} & \multicolumn{1}{l|}{RD (Trans)}         & 31.05          & 48.16          & 32.00          & \multicolumn{1}{c|}{77.62}          & 9.31           & 22.69          & 9.72           & \multicolumn{1}{c|}{58.24}          & 212.4                                                                            & 9.86                                                                              \\ \midrule
\multicolumn{12}{c}{\textbf{End-to-end DIMT (Document-level)}}                                                                                                                                                                                                                                                                                                                                                      \\ \midrule
\multicolumn{1}{l|}{14} & \multicolumn{1}{l|}{Base}               & 37.60          & 61.52          & 40.85          & \multicolumn{1}{c|}{83.08}          & 11.91          & 30.59          & 14.00          & \multicolumn{1}{c|}{52.89}          & \textbf{127.6}                                                                   & \textbf{9.16}                                                                     \\
\multicolumn{1}{l|}{15} & \multicolumn{1}{l|}{DIMTDA}             & 38.68          & 61.30          & 42.34          & \multicolumn{1}{c|}{84.44}          & 12.64          & 32.30          & 15.03          & \multicolumn{1}{c|}{\textbf{60.86}} & 242.6                                                                            & 9.82                                                                              \\
\multicolumn{1}{l|}{16} & \multicolumn{1}{l|}{M4Doc (Vary-toy)}   & 39.95          & 62.78          & 42.33          & \multicolumn{1}{c|}{83.97}          & 14.79          & 32.03          & 18.67          & \multicolumn{1}{c|}{53.73}          & 212.4                                                                            & 9.61                                                                              \\
\multicolumn{1}{l|}{17} & \multicolumn{1}{l|}{M4Doc (Vary-base)}  & 41.22          & 63.10          & 42.09          & \multicolumn{1}{c|}{86.06}          & 14.52          & 30.53          & 16.55          & \multicolumn{1}{c|}{55.89}          & 215.6                                                                            & 9.43                                                                              \\
\multicolumn{1}{l|}{18} & \multicolumn{1}{l|}{M4Doc (Llava-next)} & 34.36          & 57.88          & 37.60          & \multicolumn{1}{c|}{82.67}          & 11.03          & 30.79          & 12.58          & \multicolumn{1}{c|}{57.81}          & 216.8                                                                            & 9.96                                                                              \\
\multicolumn{1}{l|}{19} & \multicolumn{1}{l|}{M4Doc (Textmonkey)} & \textbf{42.98} & \textbf{65.41} & \textbf{44.92} & \multicolumn{1}{c|}{\textbf{86.69}} & \textbf{18.18} & \textbf{35.27} & \textbf{19.82} & \multicolumn{1}{c|}{59.98}          & 215.6                                                                            & 9.52                                                                              \\ \bottomrule

\end{tabular}
}
\caption{Results on DoTA and DITrans English-Chinese test set. The models are trained on DoTA, and tested on DoTA and DITrans. \textbf{B}, \textbf{C}, \textbf{BP}, and \textbf{S} represent BLEU, COMET, BLEU-PT, and STEDS, respectively. \textbf{\# Params} is the number of parameters of the model during inference. \textbf{Time} is the average inference time on a single NVIDIA V100 GPU. $\theta_1$ denotes the parameters of the layout analysis model and OCR model. $\theta_2$ denotes the parameters of the parameters of the layout analysis model and sentence splitting model. The \textbf{bold numbers} represent the best performance of the end-to-end DIMT.}
\label{table: main result}
\end{table*}

\subsection{Settings}
\quad \textbf{Pre-trained Models Selection } For the MLLM, we select four MLLMs with different numbers of parameters and training data: Vary-toy \citep{wei2024small}, Vary-base \citep{wei2023vary}, Llava-next \citep{liu2024llavanext} and Textmonkey \citep{liu2024textmonkey}.
The Swin Transformers of alignment encoder and image encoder are initialized from the encoder of pre-trained OCR model Nougat \citep{blecher2024nougat}.
We follow the vanilla Transformer-base \citep{vaswani2017attention} setting, pre-train an English-Chinese translation model on UN Corpus En-Zh \citep{ziemski-etal-2016-united}, and use the pre-trained decoder to initialize the translation decoder in M4Doc.

\textbf{Other Settings } The hyperparameter $\alpha$ is set to 1.0.
During training, we use the Adam optimizer and employ a linear decay learning rate schedule with a learning rate of 5e-5.
The maximum number of training steps is 15K and the batch size is 64.
More detailed settings are in Appendix~\ref{sec: appendix main experiment setting}.

\subsection{Baselines}
We evaluate our method against diverse baselines, including text-only, cascade, end-to-end, and knowledge distillation methods, to comprehensively assess its performance and validate its effectiveness.

\textbf{Text-only MT }\citep{vaswani2017attention} We use the DoTA dataset to fine-tune the Transformer-base model pre-trained on UN Corpus En-Zh \citep{ziemski-etal-2016-united}.

\subsubsection{Cascade Baselines}
\quad \textbf{LARDIT }\citep{zhang2023novel} This cascade system employs a layout analysis model \citep{yao2023docxchain}, an \href{https://github.com/tesseract-ocr/tesseract}{OCR tool}, and a text-only MT, sequentially.

\textbf{Nougat-trans } We utilize the Nougat model \citep{blecher2024nougat} for combined layout analysis and OCR and the text-only MT is employed for translation.

\subsubsection{End-to-End Baselines}
We evaluate the existing end-to-end methods under two distinct settings:  \textbf{Document-level} and \textbf{Text-line-level}.
The specific end-to-end models evaluated are:

\textbf{Base } This baseline end-to-end DIMT model uses the same image encoder and translation decoder architecture as M4Doc, without incorporating an alignment encoder or multimodal knowledge transfer.

\textbf{DIMTDA }\citep{liang-etal-2024-document} This end-to-end DIMT model uses a model assembler to integrate multiple pre-trained models to enhance the understanding of layout and translation capabilities.

\textbf{ItNet }\citep{jain2021image} This end-to-end Text Image Machine Translation (TIMT) system first pre-trains a vanilla Transformer on a text parallel dataset.
The combination of the image encoder and pre-trained decoder is fine-tuned.

\textbf{E2ETIT }\citep{ma2022improving} This end-to-end TIMT model uses a TPSNet and a ResNet as an image encoder combined with a Transformer decoder and utilizes text translation as an auxiliary task.

\textbf{PEIT }\citep{zhu-etal-2023-peit} This end-to-end TIMT system employs a vision-text representation aligner and a cross-model regularize to bridge the modality gap between visual inputs and textual inputs.

\subsubsection{Knowledge Distillation Baselines}
We conduct experiments to compare our method with three different knowledge distillation methods.

\textbf{Seq-KD }\citep{kim-rush-2016-sequence} This is the vanilla sequence-level knowledge distillation method for machine translation.

\textbf{MTKD }\citep{ma2023multi} This method employs a pre-trained OCR model and a pre-trained machine translation model as teacher models, with a TIMT model serving as the student model.

\textbf{RD }\citep{zhu-etal-2024-efficient} This approach leverages an LLM, based on the OCR results of document images, to generate rationales, subsequently employing these rationales to train a document understanding model.
As the original RD method performs poorly, we mix the generated rationales with the translation data from the DoTA dataset during training, resulting in the RD (Trans) method.

\section{Results \& Analysis}
\subsection{Main Results}
Table~\ref{table: main result} reports the performance of all methods.
It can be observed that M4Doc outperforms the cascade methods LARDIT (line 2 vs. 19) by 7.40 BLEU, 10.93 COMET and 10.86 STEDS scores on the DoTA test set.
Besides, M4Doc also achieves comparable performance with Nougat-trans (line 3 vs. 19) on both DoTA and DITrans test sets, while the number of parameters of M4Doc is reduced by 37.8\% compared to Nougat-trans and inference time decreases by 44.1\%.

Moreover, our method outperforms all the end-to-end TIMT baselines on both document-level and text-line-level settings.
As the TIMT models are designed for text-line-level images, the performances of all TIMT models under the text-line-level settings are better than document-level settings.
However, M4Doc still surpasses the highest-performing TIMT model (line 9 vs. 19) by a margin of 15.55 BLEU on the DoTA test set and 9.10 BLEU on the DITrans test set.

By comparing line 11 and line 19, our method is superior to the end-to-end DIMT baseline in in-domain and cross-domain zero-shot settings.
In the in-domain setting, there is an increase of 4.30 BLEU, 4.11 COMET, and 2.58 BLEU-PT scores.
In the cross-domain zero-shot setting, our method outperforms DIMTDA by 5.54 BLEU, 2.97 COMET, and 4.79 BLEU-PT scores, which confirms introducing MLLMs as auxiliaries during training can enhance the model's generalization abilities.

From the results presented in lines 10–13, M4Doc demonstrates superior performance compared to all knowledge distillation baselines. Furthermore, M4Doc surpasses the highest-performing baseline (lines 11 vs. 16) by 2.63 BLEU, 2.46 COMET, and 2.37 BLEU-PT scores on the DoTA test set.

From the results of line 16-19, as the number of parameters in the MLLM models increases, the DIMT model's translation quality also generally improves.
However, due to the difference in pre-training data between Llava-next and other MLLMs, MLLMs pre-trained on document images are more suitable for assisting in the training of the DIMT model.

\begin{table}[t]
\centering
\footnotesize
\resizebox{\linewidth}{!}{
\begin{tabular}{ll|cc|cc}
\toprule
\multicolumn{2}{c|}{\multirow{2}{*}{}}      & \multicolumn{2}{c|}{\textbf{Ads \& News}} & \multicolumn{2}{c}{\textbf{Political Report}} \\
\multicolumn{2}{c|}{}                       & \textbf{BLEU}                & \textbf{STEDS}              & \textbf{BLEU}                   & \textbf{STEDS}                 \\ \midrule
\multicolumn{1}{l|}{1} & DIMTDA             & 14.21               & 77.12              & 26.71                  & 89.33                 \\
\multicolumn{1}{l|}{2} & M4Doc (Vary-toy)   & 17.07               & 78.18              & 27.62                  & 88.83                 \\
\multicolumn{1}{l|}{3} & M4Doc (Vary-base)  & 21.30               & 80.67              & 31.71                  & 90.76                 \\
\multicolumn{1}{l|}{4} & M4Doc (Textmonkey) & 24.28               & 82.05              & 34.26                  & 91.06                 \\ \bottomrule
\end{tabular}
}
\caption{Results on DITrans English-Chinese test set after finetuning.}
\label{table: fine-tune results}
\end{table}

\subsection{Generalization Ability towards Difficult DIMT Scenarios}
We pay special attention to challenging DIMT scenarios, where our model exhibits advantages through single-to-mix modality alignment.
Consequently, we conduct three sets of experiments.
\subsubsection{Cross-domain}
We fine-tune end-to-end DIMT models on two subsets of the DITrans dataset separately after training on the DoTA dataset.
Detailed dataset setting can be seen in Appendix~\ref{sec: appendix dataset setting}.
The results are shown in Table~\ref{table: fine-tune results}.
With the help of MLLM during training, all three variants of our method achieve better performance than DIMTDA.
This could be because the MLLM is pre-trained on a large amount of data, allowing the alignment encoder to learn similar representations from the MLLM, thus enhancing the generalization capability of the DIMT model.

\begin{figure}[t]
    \centering
    \includegraphics[width=\columnwidth]{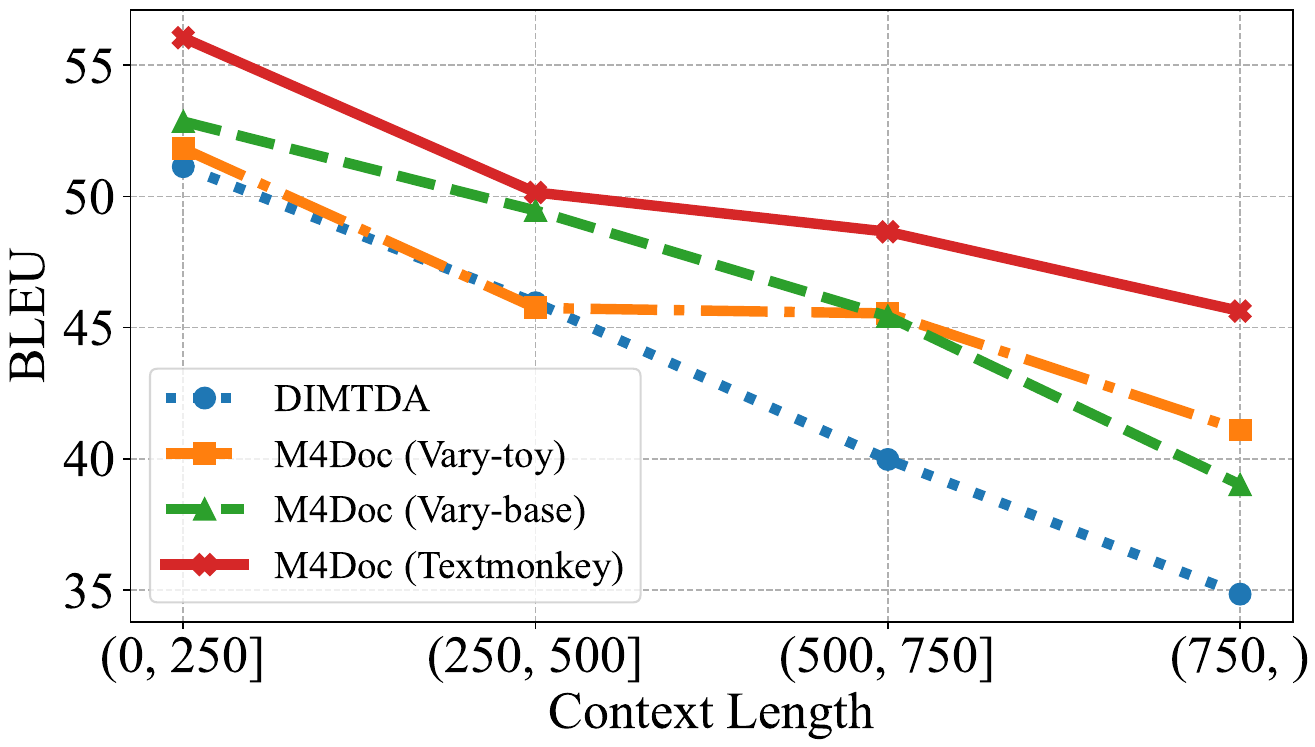}
    \caption{BLEU scores of M4Doc models testing on different context length valid sets. Detailed data can be seen in Appendix~\ref{sec: appendix detailed data}.}
    \label{figure: context length}
\end{figure}

\subsubsection{Long Context}
We select samples from the valid set within different context lengths.\footnote{Context length refers to the number of English words in the image.}
Detailed settings can be seen in Appendix~\ref{sec: appendix dataset setting}.
Results are shown in Figure~\ref{figure: context length}.
Our models outperform the baseline across all context length scenarios.
The performance of all models decreases as the context length increases, but the decline is less pronounced with our models compared to DIMTDA, which indicates that the introducing of MLLM can improve the DIMT models' ability to handle images with long context.

\subsubsection{Complex Layout}
\label{sec: layout complexity}
We select two subsets (images with simple layout and complex layout) from the valid set of the DoTA dataset.
Detailed settings can be seen in Appendix~\ref{sec: appendix dataset setting}.
As shown in Table~\ref{table: layout complexity}, our methods perform similarly to DIMTDA on images with simple layouts, but on images with complex layouts, our methods can achieve up to 5.58 BLEU and 6.08 BLEU-PT scores higher than DIMTDA.
It suggests that the assistance of the MLLM during training can improve the DIMT models' ability to understand complex layout structures and further improve the translation ability.

\begin{table}[t]
\footnotesize
\centering
\resizebox{\linewidth}{!}{
\begin{tabular}{lcl|ccc}
\toprule
\multicolumn{3}{c|}{}                                                                  & \textbf{BLEU}  & \textbf{BLEU-PT} & \textbf{STEDS} \\ \midrule
\multicolumn{1}{l|}{1} & \multicolumn{1}{c|}{\multirow{2}{*}{Simple}}  & DIMTDA & 55.24 & 55.26   & 90.54 \\
\multicolumn{1}{l|}{2} & \multicolumn{1}{c|}{}                                & M4Doc  & 56.88 & 56.72   & 92.25 \\ \midrule
\multicolumn{1}{l|}{3} & \multicolumn{1}{c|}{\multirow{2}{*}{Complex}} & DIMTDA & 30.30 & 35.16   & 84.57 \\
\multicolumn{1}{l|}{4} & \multicolumn{1}{c|}{}                                & M4Doc  & 35.88 & 41.24   & 83.76 \\ \bottomrule
\end{tabular}
}
\caption{Results of different layout complexity on DoTA English-Chinese valid set.}
\label{table: layout complexity}
\end{table}

\begin{table}[t]
\footnotesize
\centering
\resizebox{\linewidth}{!}{
\begin{tabular}{ll|ccc}
\toprule
\multicolumn{2}{c|}{}                           & \textbf{BLEU}  & \textbf{BLEU-PT} & \textbf{STEDS} \\ \midrule
\multicolumn{1}{l|}{1} & M4Doc (Vary-toy)       & 40.05 & 42.58   & 83.93 \\ \midrule
\multicolumn{1}{l|}{2} & w/o $\mathcal{L}_{\mathrm{align}}$        & 36.58 & 40.06   & 83.54 \\
\multicolumn{1}{l|}{3} & w/o Alignment Encoder   & 36.62 & 40.09   & 83.65 \\
\multicolumn{1}{l|}{4} & w MLLM Output          & 42.56 & 46.93   & 89.48 \\ \bottomrule
\end{tabular}
}
\caption{Ablation study results on DoTA English-Chinese valid set.}
\label{table: ablation study results}
\end{table}

\subsection{Ablation Study}
\subsubsection{Effect of Different Module}
To investigate the effectiveness of the proposed modules, we conduct ablation experiments.
The results are shown in Table~\ref{table: ablation study results}.

\textbf{w/o $\mathcal{L}_{\mathrm{align}}$ } We remove the MLLM during training, keep the alignment encoder, and only use $\mathcal{L}_{\mathrm{trans}}$ to guide the model.
By comparing line 1 and line 2, a decline of 3.47 BLEU and 2.52 BLEU-PT scores can be observed, which demonstrates the effectiveness of the MLLM during training.

\textbf{w/o Alignment Encoder } We remove the Swin Transformer in the alignment encoder and use the output of the image encoder to do alignment with MLLM and image encoding simultaneously.
It can be seen from the comparison between line 1 and line 3 that simultaneously achieving alignment and image encoding is challenging for a single encoder and causes a decrease in translation quality.

\textbf{w MLLM Output } The output hidden states of MLLM are directly sent to the translation decoder without the alignment encoder as an intermediary.
By comparing line 1 and line 4, there is an increase of 2.51 BLEU and 5.55 STEDS scores.
However, this approach significantly increases the parameters of the model ($\times$ 11.53) and inference time ($\times$ 1.26).
Our method strikes a balance between translation quality and inference speed.

\begin{table}[t]
\footnotesize
\centering
\resizebox{\linewidth}{!}{
\begin{tabular}{ll|ccc}
\toprule
\multicolumn{2}{c|}{}                           & \textbf{BLEU}  & \textbf{BLEU-PT} & \textbf{STEDS} \\ \midrule
\multicolumn{1}{l|}{1} & M4Doc (Vary-toy)       & 40.05 & 42.58   & 83.93 \\ \midrule
\multicolumn{1}{l|}{2} & w/o MLLM Image Input   & 38.77 & 42.00   & 84.99 \\
\multicolumn{1}{l|}{3} & w/o MLLM Text Input    & 37.06 & 38.69   & 84.98 \\ \bottomrule
\end{tabular}
}
\caption{Results on DoTA English-Chinese valid set with different modalities input.}
\label{table: mllm input}
\end{table}

\begin{figure}[t]
    \centering
    \includegraphics[width=\columnwidth]{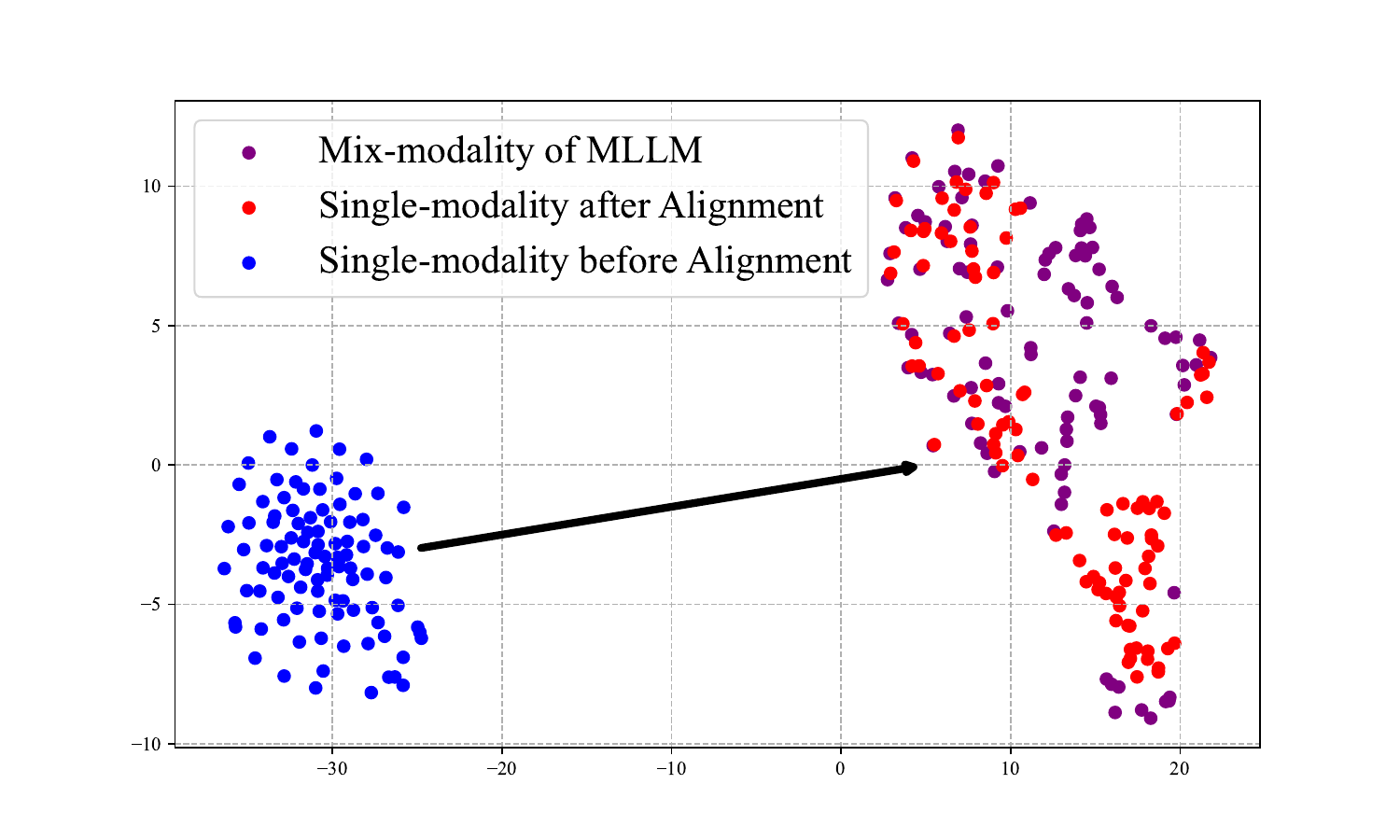}
    \caption{T-SNE visualization of different representations for MLLM and alignment encoder.}
    \label{figure: tsne}
\end{figure}

\subsubsection{Effect of Mix-modality Input}
To explore the impact of mix-modality input, we only send English text or the corresponding image to the MLLM during training.
The input formats are \texttt{<System Prompt> <Image Token>} and \texttt{<System Prompt> <Source Text>}.
As the results of Table~\ref{table: mllm input} show, the performance degradation of the model is greater when text input is removed compared to when image input is removed.
This may be because the source text contains more translation-relevant textual information.
When MLLM has only image input, the alignment from image modality to image modality does not introduce additional information.

We provide a visualization of the representations of 100 samples in Figure~\ref{figure: tsne}.
The single-modality representation output by the alignment encoder, after training, largely overlaps with the distribution of the mix-modality representation output by the MLLM.
This demonstrates that our proposed single-to-mix modality alignment allows the alignment encoder to effectively learn the MLLM outputs, providing additional information to guide the translation decoder in generating translations.

\begin{figure}[t]
    \centering
    \includegraphics[width=\columnwidth]{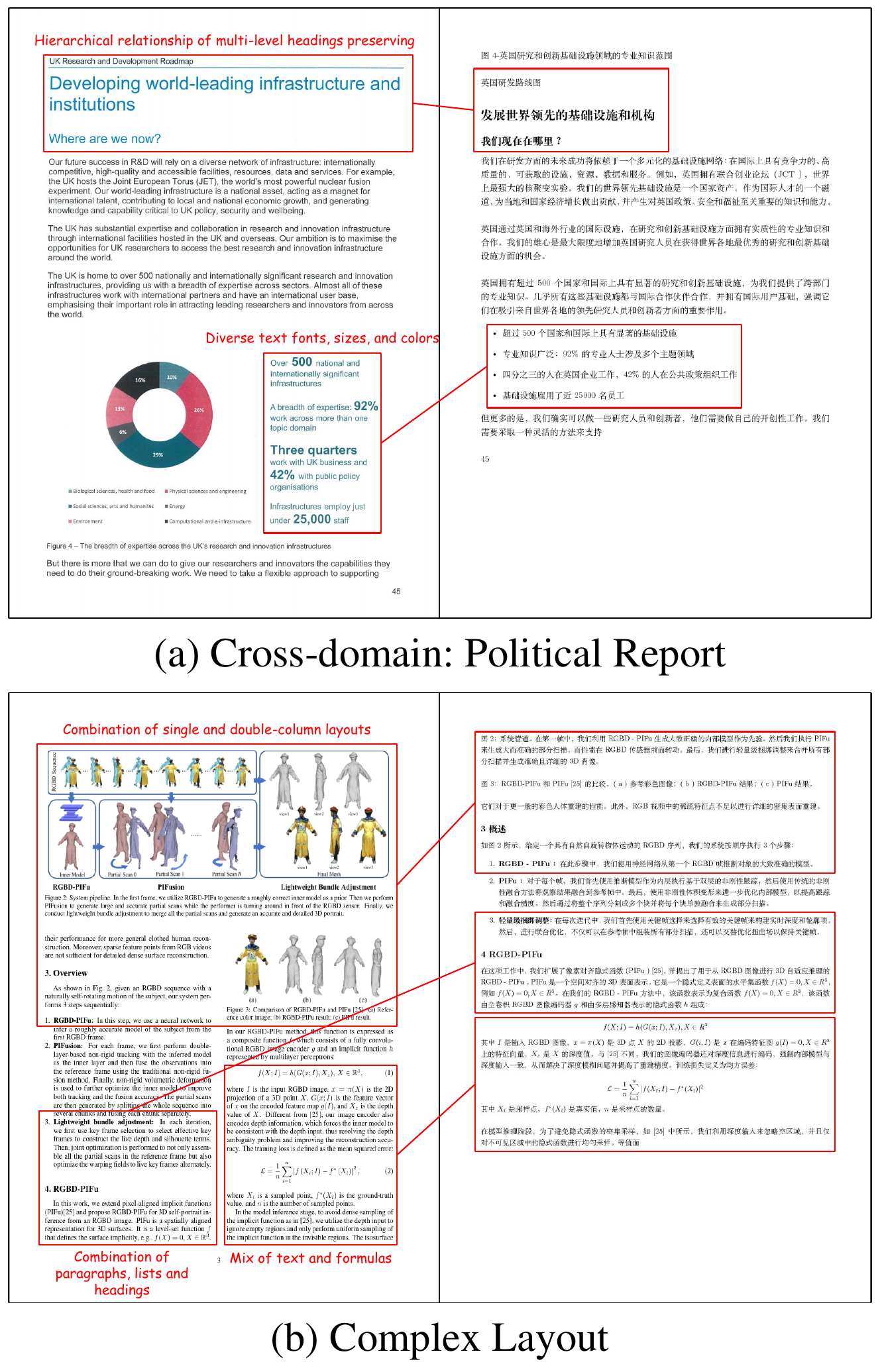}
    \caption{The output samples of M4Doc. For each image pair, the left is the input document image, and the right is the output translations in markdown format after rendering. It is better to zoom in for a clearer view. More samples can be seen in Appendix~\ref{sec: appendix output samples}. }
    \label{figure: case study}
\end{figure}

\subsection{Case Study}
We provide the output samples of M4Doc cross-domain and complex layout scenarios in Figure~\ref{figure: case study}.
More samples can be seen in Appendix~\ref{sec: appendix output samples}.

Figure~\ref{figure: case study} (a) is an image from the political report subset of the DITrans dataset.
The fonts, sizes, and colors of the texts are diverse, which is quite different from the DoTA dataset used for training.
After fine-tuning, our model can still perform translations and obtain the hierarchical relationship of multi-level headings and lists.

Figure~\ref{figure: case study} (b) comes from the DoTA dataset.
The image contains a mix of text, figures, and formulas, with headings and lists, and a combination of single and double-column layouts, resulting in a highly complex layout structure.
Our model can still output the translation results in a logical order, formatted in Markdown.

\section{Related Work}
Text Image Machine Translation (TIMT) refers to translating texts from one language to another within images, as explored by \citet{lan-etal-2023-exploring}.
In recent years, various end-to-end TIMT methods \citep{ma2023modal, ma2023e2timt, ma2023multi, ma-etal-2024-born, tian-etal-2023-image, ma-etal-2024-born-babynet, lan2024translatotron, qian2024anytrans, guan-etal-2025-trifine} have been proposed.
\citet{jain2021image} follows the encoder-decoder paradigm and uses a convolutional encoder and an autoregressive Transformer decoder to build the model.
\citet{ma2022improving} proposes a text translation enhanced text image translation method, which trains the end-to-end TIMT model with text translation as an auxiliary task.
\citet{zhu-etal-2023-peit} introduces an end-to-end TIMT framework that bridges the modality gap with pre-trained models.
While these end-to-end methods have demonstrated satisfactory performance, their effectiveness is limited to images with short context and simple layout structure, different from document images.

Recent advancements in MLLMs have significantly improved the processing and understanding of text-rich images \citep{ye2023mplug, wei2024small, hu2024mplug, yu2024texthawk}.
\citet{wei2023vary} explores adding fine-grained vision perception for document images to the MLLM without affecting its existing natural image understanding capabilities.
\citet{zhang2023llavar} and \citet{liu2024llavanext} utilize GPT-4 \citep{yang2023dawn} to construct a visual instruction tuning dataset and improve LLaVA's \citep{liu2023visual} ability to comprehend textual detail within images.
\citet{liu2024textmonkey} proposes shifted window attention to achieve cross-window connectivity at higher input resolutions and token resampler to filter out significant tokens.
As MLLMs take both images and texts as inputs during the pre-training stage, the integration of visual and linguistic information provides a better understanding of document images, which inspires us to leverage the MLLM for the DIMT task.

\section{Conclusion}
In this paper, we propose a novel method, single-to-mix modality alignment with multimodal large language model for document image machine translation (M4Doc), which has three advantages.
Firstly, single-to-mix modality alignment allows the alignment encoder to infer more textual information from the image input.
Secondly, the alignment with MLLM enhances the generalization capability towards three difficult DIMT scenarios.
Finally, the introduction of the alignment encoder achieves the SOTA translation quality while preserving high inference efficiency.
Extensive experiments demonstrate the effectiveness of M4Doc and highlight its advantage in enhancing the performance of the DIMT model in cross-domain and complex document image scenarios.

\section*{Limitations}
Although M4Doc achieves notable results on the DIMT task, current end-to-end models generate the entire translated text of the document image in a single output.
In the future, we plan to explore integrating user prompts to translate text in specific regions of the image, thereby making the translation more aligned with user preferences.

\section*{Acknowledgements}
We thank anonymous reviewers for helpful suggestions.
This work is supported by the National Natural Science Foundation of China (No. 62336008 and No. 62476275).

\bibliography{acl_latex}

\begin{thebibliography}{42}
\providecommand{\natexlab}[1]{#1}

\bibitem[{Blecher et~al.(2024)Blecher, Cucurull, Scialom, and Stojnic}]{blecher2024nougat}
Lukas Blecher, Guillem Cucurull, Thomas Scialom, and Robert Stojnic. 2024.
\newblock \href {https://openreview.net/forum?id=fUtxNAKpdV} {Nougat: Neural optical understanding for academic documents}.
\newblock In \emph{The Twelfth International Conference on Learning Representations}.

\bibitem[{Guan et~al.(2025)Guan, Zhang, Zhao, and Zong}]{guan-etal-2025-trifine}
Boyu Guan, Yining Zhang, Yang Zhao, and Chengqing Zong. 2025.
\newblock \href {https://aclanthology.org/2025.coling-main.547/} {{T}ri{F}ine: A large-scale dataset of vision-audio-subtitle for tri-modal machine translation and benchmark with fine-grained annotated tags}.
\newblock In \emph{Proceedings of the 31st International Conference on Computational Linguistics}, pages 8215--8231, Abu Dhabi, UAE. Association for Computational Linguistics.

\bibitem[{Hinami et~al.(2021)Hinami, Ishiwatari, Yasuda, and Matsui}]{hinami2021towards}
Ryota Hinami, Shonosuke Ishiwatari, Kazuhiko Yasuda, and Yusuke Matsui. 2021.
\newblock Towards fully automated manga translation.
\newblock In \emph{Proceedings of the AAAI Conference on Artificial Intelligence}, volume~35, pages 12998--13008.

\bibitem[{Hu et~al.(2024)Hu, Xu, Ye, Yan, Zhang, Zhang, Li, Zhang, Jin, Huang et~al.}]{hu2024mplug}
Anwen Hu, Haiyang Xu, Jiabo Ye, Ming Yan, Liang Zhang, Bo~Zhang, Chen Li, Ji~Zhang, Qin Jin, Fei Huang, et~al. 2024.
\newblock mplug-docowl 1.5: Unified structure learning for ocr-free document understanding.
\newblock \emph{arXiv preprint arXiv:2403.12895}.

\bibitem[{Hu et~al.(2022)Hu, Shen, Wallis, Allen{-}Zhu, Li, Wang, Wang, and Chen}]{DBLP:conf/iclr/HuSWALWWC22}
Edward~J. Hu, Yelong Shen, Phillip Wallis, Zeyuan Allen{-}Zhu, Yuanzhi Li, Shean Wang, Lu~Wang, and Weizhu Chen. 2022.
\newblock \href {https://openreview.net/forum?id=nZeVKeeFYf9} {Lora: Low-rank adaptation of large language models}.
\newblock In \emph{The Tenth International Conference on Learning Representations, {ICLR} 2022, Virtual Event, April 25-29, 2022}. OpenReview.net.

\bibitem[{Hurst et~al.(2024)Hurst, Lerer, Goucher, Perelman, Ramesh, Clark, Ostrow, Welihinda, Hayes, Radford et~al.}]{hurst2024gpt}
Aaron Hurst, Adam Lerer, Adam~P Goucher, Adam Perelman, Aditya Ramesh, Aidan Clark, AJ~Ostrow, Akila Welihinda, Alan Hayes, Alec Radford, et~al. 2024.
\newblock Gpt-4o system card.
\newblock \emph{arXiv preprint arXiv:2410.21276}.

\bibitem[{Jain et~al.(2021)Jain, Firat, Ge, and Liang}]{jain2021image}
Puneet Jain, Orhan Firat, Qi~Ge, and Sihang Liang. 2021.
\newblock Image translation network.

\bibitem[{Kim and Rush(2016)}]{kim-rush-2016-sequence}
Yoon Kim and Alexander~M. Rush. 2016.
\newblock \href {https://doi.org/10.18653/v1/D16-1139} {Sequence-level knowledge distillation}.
\newblock In \emph{Proceedings of the 2016 Conference on Empirical Methods in Natural Language Processing}, pages 1317--1327, Austin, Texas. Association for Computational Linguistics.

\bibitem[{Lan et~al.(2024)Lan, Niu, Meng, Zhou, Zhang, and Su}]{lan2024translatotron}
Zhibin Lan, Liqiang Niu, Fandong Meng, Jie Zhou, Min Zhang, and Jinsong Su. 2024.
\newblock Translatotron-v (ison): An end-to-end model for in-image machine translation.
\newblock \emph{arXiv preprint arXiv:2407.02894}.

\bibitem[{Lan et~al.(2023)Lan, Yu, Li, Zhang, Luan, Wang, Huang, and Su}]{lan-etal-2023-exploring}
Zhibin Lan, Jiawei Yu, Xiang Li, Wen Zhang, Jian Luan, Bin Wang, Degen Huang, and Jinsong Su. 2023.
\newblock \href {https://doi.org/10.18653/v1/2023.acl-long.192} {Exploring better text image translation with multimodal codebook}.
\newblock In \emph{Proceedings of the 61st Annual Meeting of the Association for Computational Linguistics (Volume 1: Long Papers)}, pages 3479--3491, Toronto, Canada. Association for Computational Linguistics.

\bibitem[{Liang et~al.(2024)Liang, Zhang, Ma, Zhang, Zhao, Xiang, Zong, and Zhou}]{liang-etal-2024-document}
Yupu Liang, Yaping Zhang, Cong Ma, Zhiyang Zhang, Yang Zhao, Lu~Xiang, Chengqing Zong, and Yu~Zhou. 2024.
\newblock \href {https://aclanthology.org/2024.naacl-long.392} {Document image machine translation with dynamic multi-pre-trained models assembling}.
\newblock In \emph{Proceedings of the 2024 Conference of the North American Chapter of the Association for Computational Linguistics: Human Language Technologies (Volume 1: Long Papers)}, pages 7084--7095, Mexico City, Mexico. Association for Computational Linguistics.

\bibitem[{Liu et~al.(2024{\natexlab{a}})Liu, Li, Li, Li, Zhang, Shen, and Lee}]{liu2024llavanext}
Haotian Liu, Chunyuan Li, Yuheng Li, Bo~Li, Yuanhan Zhang, Sheng Shen, and Yong~Jae Lee. 2024{\natexlab{a}}.
\newblock \href {https://llava-vl.github.io/blog/2024-01-30-llava-next/} {Llava-next: Improved reasoning, ocr, and world knowledge}.

\bibitem[{Liu et~al.(2023)Liu, Li, Wu, and Lee}]{liu2023visual}
Haotian Liu, Chunyuan Li, Qingyang Wu, and Yong~Jae Lee. 2023.
\newblock Visual instruction tuning.
\newblock In \emph{Proceedings of the 37th International Conference on Neural Information Processing Systems}, pages 34892--34916.

\bibitem[{Liu et~al.(2024{\natexlab{b}})Liu, Yang, Liu, Li, Ma, Zhang, and Bai}]{liu2024textmonkey}
Yuliang Liu, Biao Yang, Qiang Liu, Zhang Li, Zhiyin Ma, Shuo Zhang, and Xiang Bai. 2024{\natexlab{b}}.
\newblock Textmonkey: An ocr-free large multimodal model for understanding document.
\newblock \emph{arXiv preprint arXiv:2403.04473}.

\bibitem[{Ma et~al.(2023{\natexlab{a}})Ma, Han, Wu, Zhang, Zhao, Zhou, and Zong}]{ma2023modal}
Cong Ma, Xu~Han, Linghui Wu, Yaping Zhang, Yang Zhao, Yu~Zhou, and Chengqing Zong. 2023{\natexlab{a}}.
\newblock Modal contrastive learning based end-to-end text image machine translation.
\newblock \emph{IEEE/ACM Transactions on Audio, Speech, and Language Processing}.

\bibitem[{Ma et~al.(2022)Ma, Zhang, Tu, Han, Wu, Zhao, and Zhou}]{ma2022improving}
Cong Ma, Yaping Zhang, Mei Tu, Xu~Han, Linghui Wu, Yang Zhao, and Yu~Zhou. 2022.
\newblock Improving end-to-end text image translation from the auxiliary text translation task.
\newblock In \emph{2022 26th International Conference on Pattern Recognition (ICPR)}, pages 1664--1670. IEEE.

\bibitem[{Ma et~al.(2023{\natexlab{b}})Ma, Zhang, Tu, Zhao, Zhou, and Zong}]{ma2023e2timt}
Cong Ma, Yaping Zhang, Mei Tu, Yang Zhao, Yu~Zhou, and Chengqing Zong. 2023{\natexlab{b}}.
\newblock E2timt: Efficient and effective modal adapter for text image machine translation.
\newblock In \emph{The 17th International Conference on Document Analysis and Recognition (ICDAR)}, pages 70--88.

\bibitem[{Ma et~al.(2023{\natexlab{c}})Ma, Zhang, Tu, Zhao, Zhou, and Zong}]{ma2023multi}
Cong Ma, Yaping Zhang, Mei Tu, Yang Zhao, Yu~Zhou, and Chengqing Zong. 2023{\natexlab{c}}.
\newblock Multi-teacher knowledge distillation for text image machine translation.
\newblock In \emph{The 17th International Conference on Document Analysis and Recognition (ICDAR)}, pages 484--501.

\bibitem[{Ma et~al.(2024{\natexlab{a}})Ma, Zhang, Zhang, Liang, Zhao, Zhou, and Zong}]{ma-etal-2024-born}
Cong Ma, Yaping Zhang, Zhiyang Zhang, Yupu Liang, Yang Zhao, Yu~Zhou, and Chengqing Zong. 2024{\natexlab{a}}.
\newblock \href {https://aclanthology.org/2024.lrec-main.222/} {Born a {B}aby{N}et with hierarchical parental supervision for end-to-end text image machine translation}.
\newblock In \emph{Proceedings of the 2024 Joint International Conference on Computational Linguistics, Language Resources and Evaluation (LREC-COLING 2024)}, pages 2468--2479, Torino, Italia. ELRA and ICCL.

\bibitem[{Ma et~al.(2024{\natexlab{b}})Ma, Zhang, Zhang, Liang, Zhao, Zhou, and Zong}]{ma-etal-2024-born-babynet}
Cong Ma, Yaping Zhang, Zhiyang Zhang, Yupu Liang, Yang Zhao, Yu~Zhou, and Chengqing Zong. 2024{\natexlab{b}}.
\newblock \href {https://aclanthology.org/2024.lrec-main.222} {Born a {B}aby{N}et with hierarchical parental supervision for end-to-end text image machine translation}.
\newblock In \emph{Proceedings of the 2024 Joint International Conference on Computational Linguistics, Language Resources and Evaluation (LREC-COLING 2024)}, pages 2468--2479, Torino, Italia. ELRA and ICCL.

\bibitem[{Nguyen et~al.(2021)Nguyen, Lai, Pouran Ben~Veyseh, and Nguyen}]{nguyen-etal-2021-trankit}
Minh~Van Nguyen, Viet~Dac Lai, Amir Pouran Ben~Veyseh, and Thien~Huu Nguyen. 2021.
\newblock \href {https://doi.org/10.18653/v1/2021.eacl-demos.10} {Trankit: A light-weight transformer-based toolkit for multilingual natural language processing}.
\newblock In \emph{Proceedings of the 16th Conference of the European Chapter of the Association for Computational Linguistics: System Demonstrations}, pages 80--90, Online. Association for Computational Linguistics.

\bibitem[{Qian et~al.(2024)Qian, Zhang, Yang, Fan, Ma, Wong, Sun, and Ji}]{qian2024anytrans}
Zhipeng Qian, Pei Zhang, Baosong Yang, Kai Fan, Yiwei Ma, Derek~F Wong, Xiaoshuai Sun, and Rongrong Ji. 2024.
\newblock Anytrans: Translate anytext in the image with large scale models.
\newblock \emph{arXiv preprint arXiv:2406.11432}.

\bibitem[{Rei et~al.(2020)Rei, Stewart, Farinha, and Lavie}]{rei-etal-2020-comet}
Ricardo Rei, Craig Stewart, Ana~C Farinha, and Alon Lavie. 2020.
\newblock \href {https://doi.org/10.18653/v1/2020.emnlp-main.213} {{COMET}: A neural framework for {MT} evaluation}.
\newblock In \emph{Proceedings of the 2020 Conference on Empirical Methods in Natural Language Processing (EMNLP)}, pages 2685--2702, Online. Association for Computational Linguistics.

\bibitem[{Sable et~al.(2023)Sable, Shelke, Deogaonkar, Joshi, Kabadi, and Joshi}]{sable2023doc}
Nilesh~P Sable, Priya Shelke, Ninad Deogaonkar, Nachiket Joshi, Rudra Kabadi, and Tushar Joshi. 2023.
\newblock Doc-handler: Document scanner, manipulator, and translator based on image and natural language processing.
\newblock In \emph{2023 International Conference on Emerging Smart Computing and Informatics (ESCI)}, pages 1--6. IEEE.

\bibitem[{Steingrimsson et~al.(2023)Steingrimsson, Loftsson, and Way}]{steingrimsson-etal-2023-sentalign}
Steinthor Steingrimsson, Hrafn Loftsson, and Andy Way. 2023.
\newblock \href {https://doi.org/10.18653/v1/2023.emnlp-demo.22} {{S}ent{A}lign: Accurate and scalable sentence alignment}.
\newblock In \emph{Proceedings of the 2023 Conference on Empirical Methods in Natural Language Processing: System Demonstrations}, pages 256--263, Singapore. Association for Computational Linguistics.

\bibitem[{Team et~al.(2024)Team, Georgiev, Lei, Burnell, Bai, Gulati, Tanzer, Vincent, Pan, Wang et~al.}]{team2024gemini}
Gemini Team, Petko Georgiev, Ving~Ian Lei, Ryan Burnell, Libin Bai, Anmol Gulati, Garrett Tanzer, Damien Vincent, Zhufeng Pan, Shibo Wang, et~al. 2024.
\newblock Gemini 1.5: Unlocking multimodal understanding across millions of tokens of context.
\newblock \emph{arXiv preprint arXiv:2403.05530}.

\bibitem[{Tian et~al.(2023)Tian, Li, Liu, Guo, and Wang}]{tian-etal-2023-image}
Yanzhi Tian, Xiang Li, Zeming Liu, Yuhang Guo, and Bin Wang. 2023.
\newblock \href {https://doi.org/10.18653/v1/2023.findings-emnlp.1004} {In-image neural machine translation with segmented pixel sequence-to-sequence model}.
\newblock In \emph{Findings of the Association for Computational Linguistics: EMNLP 2023}, pages 15046--15057, Singapore. Association for Computational Linguistics.

\bibitem[{Vaswani et~al.(2017)Vaswani, Shazeer, Parmar, Uszkoreit, Jones, Gomez, Kaiser, and Polosukhin}]{vaswani2017attention}
Ashish Vaswani, Noam Shazeer, Niki Parmar, Jakob Uszkoreit, Llion Jones, Aidan~N Gomez, {\L}ukasz Kaiser, and Illia Polosukhin. 2017.
\newblock Attention is all you need.
\newblock \emph{Advances in neural information processing systems}, 30.

\bibitem[{Wei et~al.(2023)Wei, Kong, Chen, Zhao, Ge, Yang, Sun, Han, and Zhang}]{wei2023vary}
Haoran Wei, Lingyu Kong, Jinyue Chen, Liang Zhao, Zheng Ge, Jinrong Yang, Jianjian Sun, Chunrui Han, and Xiangyu Zhang. 2023.
\newblock Vary: Scaling up the vision vocabulary for large vision-language models.
\newblock \emph{arXiv preprint arXiv:2312.06109}.

\bibitem[{Wei et~al.(2024)Wei, Kong, Chen, Zhao, Ge, Yu, Sun, Han, and Zhang}]{wei2024small}
Haoran Wei, Lingyu Kong, Jinyue Chen, Liang Zhao, Zheng Ge, En~Yu, Jianjian Sun, Chunrui Han, and Xiangyu Zhang. 2024.
\newblock Small language model meets with reinforced vision vocabulary.
\newblock \emph{arXiv preprint arXiv:2401.12503}.

\bibitem[{Yang et~al.(2023)Yang, Li, Lin, Wang, Lin, Liu, and Wang}]{yang2023dawn}
Zhengyuan Yang, Linjie Li, Kevin Lin, Jianfeng Wang, Chung-Ching Lin, Zicheng Liu, and Lijuan Wang. 2023.
\newblock The dawn of lmms: Preliminary explorations with gpt-4v (ision).
\newblock \emph{arXiv preprint arXiv:2309.17421}, 9(1):1.

\bibitem[{Yao(2023)}]{yao2023docxchain}
Cong Yao. 2023.
\newblock Docxchain: A powerful open-source toolchain for document parsing and beyond.
\newblock \emph{arXiv preprint arXiv:2310.12430}.

\bibitem[{Ye et~al.(2023)Ye, Hu, Xu, Ye, Yan, Dan, Zhao, Xu, Li, Tian et~al.}]{ye2023mplug}
Jiabo Ye, Anwen Hu, Haiyang Xu, Qinghao Ye, Ming Yan, Yuhao Dan, Chenlin Zhao, Guohai Xu, Chenliang Li, Junfeng Tian, et~al. 2023.
\newblock mplug-docowl: Modularized multimodal large language model for document understanding.
\newblock \emph{arXiv preprint arXiv:2307.02499}.

\bibitem[{Yu et~al.(2024)Yu, Liao, Wu, Liao, Zheng, and Zeng}]{yu2024texthawk}
Ya-Qi Yu, Minghui Liao, Jihao Wu, Yongxin Liao, Xiaoyu Zheng, and Wei Zeng. 2024.
\newblock Texthawk: Exploring efficient fine-grained perception of multimodal large language models.
\newblock \emph{arXiv preprint arXiv:2404.09204}.

\bibitem[{Zhang et~al.(2023{\natexlab{a}})Zhang, Zhang, Gu, Zhou, Lipka, Yang, and Sun}]{zhang2023llavar}
Yanzhe Zhang, Ruiyi Zhang, Jiuxiang Gu, Yufan Zhou, Nedim Lipka, Diyi Yang, and Tong Sun. 2023{\natexlab{a}}.
\newblock Llavar: Enhanced visual instruction tuning for text-rich image understanding.
\newblock \emph{arXiv preprint arXiv:2306.17107}.

\bibitem[{Zhang et~al.(2025{\natexlab{a}})Zhang, Zhang, Liang, Ma, Xiang, Zhao, Zhou, and Zong}]{DBLP:journals/pami/ZhangZLMXZZZ25}
Zhiyang Zhang, Yaping Zhang, Yupu Liang, Cong Ma, Lu~Xiang, Yang Zhao, Yu~Zhou, and Chengqing Zong. 2025{\natexlab{a}}.
\newblock \href {https://doi.org/10.1109/TPAMI.2025.3530998} {Understand layout and translate text: Unified feature-conductive end-to-end document image translation}.
\newblock \emph{{IEEE} Trans. Pattern Anal. Mach. Intell.}, 47(5):3358--3376.

\bibitem[{Zhang et~al.(2023{\natexlab{b}})Zhang, Zhang, Liang, Xiang, Zhao, Zhou, and Zong}]{zhang-etal-2023-layoutdit}
Zhiyang Zhang, Yaping Zhang, Yupu Liang, Lu~Xiang, Yang Zhao, Yu~Zhou, and Chengqing Zong. 2023{\natexlab{b}}.
\newblock \href {https://doi.org/10.18653/v1/2023.findings-emnlp.673} {{L}ayout{DIT}: Layout-aware end-to-end document image translation with multi-step conductive decoder}.
\newblock In \emph{Findings of the Association for Computational Linguistics: EMNLP 2023}, pages 10043--10053, Singapore. Association for Computational Linguistics.

\bibitem[{Zhang et~al.(2025{\natexlab{b}})Zhang, Zhang, Liang, Xiang, Zhao, Zhou, and Zong}]{zhang-etal-2025-chaotic}
Zhiyang Zhang, Yaping Zhang, Yupu Liang, Lu~Xiang, Yang Zhao, Yu~Zhou, and Chengqing Zong. 2025{\natexlab{b}}.
\newblock \href {https://aclanthology.org/2025.coling-main.723/} {From chaotic {OCR} words to coherent document: A fine-to-coarse zoom-out network for complex-layout document image translation}.
\newblock In \emph{Proceedings of the 31st International Conference on Computational Linguistics}, pages 10877--10890, Abu Dhabi, UAE. Association for Computational Linguistics.

\bibitem[{Zhang et~al.(2023{\natexlab{c}})Zhang, Zhang, Xiang, Zhao, Zhou, and Zong}]{zhang2023novel}
Zhiyang Zhang, Yaping Zhang, Lu~Xiang, Yang Zhao, Yu~Zhou, and Chengqing Zong. 2023{\natexlab{c}}.
\newblock A novel dataset and benchmark analysis on document image translation.
\newblock In \emph{China Conference on Machine Translation}, pages 103--115. Springer.

\bibitem[{Zhu et~al.(2023)Zhu, Li, Lei, and Xiong}]{zhu-etal-2023-peit}
Shaolin Zhu, Shangjie Li, Yikun Lei, and Deyi Xiong. 2023.
\newblock \href {https://doi.org/10.18653/v1/2023.acl-long.751} {{PEIT}: Bridging the modality gap with pre-trained models for end-to-end image translation}.
\newblock In \emph{Proceedings of the 61st Annual Meeting of the Association for Computational Linguistics (Volume 1: Long Papers)}, pages 13433--13447, Toronto, Canada. Association for Computational Linguistics.

\bibitem[{Zhu et~al.(2024)Zhu, Agarwal, Joshi, Jia, Thomason, and Toutanova}]{zhu-etal-2024-efficient}
Wang Zhu, Alekh Agarwal, Mandar Joshi, Robin Jia, Jesse Thomason, and Kristina Toutanova. 2024.
\newblock \href {https://doi.org/10.18653/v1/2024.naacl-long.465} {Efficient end-to-end visual document understanding with rationale distillation}.
\newblock In \emph{Proceedings of the 2024 Conference of the North American Chapter of the Association for Computational Linguistics: Human Language Technologies (Volume 1: Long Papers)}, pages 8401--8424, Mexico City, Mexico. Association for Computational Linguistics.

\bibitem[{Ziemski et~al.(2016)Ziemski, Junczys-Dowmunt, and Pouliquen}]{ziemski-etal-2016-united}
Micha{\l} Ziemski, Marcin Junczys-Dowmunt, and Bruno Pouliquen. 2016.
\newblock \href {https://aclanthology.org/L16-1561} {The {U}nited {N}ations parallel corpus v1.0}.
\newblock In \emph{Proceedings of the Tenth International Conference on Language Resources and Evaluation ({LREC}'16)}, pages 3530--3534, Portoro{\v{z}}, Slovenia. European Language Resources Association (ELRA).

\end{thebibliography}


\appendix
\section*{Appendix}
\section{Setting Details}
\subsection{Dataset Settings}
\label{sec: appendix dataset setting}
In the DITrans dataset, the sample sizes for the advertisement, news, and political report subdomains are 485, 610, and 1397, respectively.
Due to the small number of images in the advertisement and news domains and their similar layout structures as scanned document images, we merge these two domains.
We then randomly select 100 images as the test set and another 100 images as the valid set.
For the political report domain, we also randomly select 100 images as the test set and another 100 images as the valid set.

In the experiment of varying context length, to shield the impact of layout difference, we select images with a single column and without formulas, tables, or figures.
Context length refers to the number of English words in the image.

In the experiment of varying layout complexity, we transform samples from the valid set into trees, selecting the 100 trees with the fewest nodes as a simple layout set and the 100 trees with the most nodes as a complex layout set.



\subsection{Main Experiment Settings}
\label{sec: appendix main experiment setting}
We segment the Chinese texts with \href{https://github.com/fxsjy/jieba}{jieba} and apply WordPiece to segment both English and Chinese texts and the vocabulary size of both English and Chinese is 52K.
We use the pre-trained OCR model Nougat’s encoder \citep{blecher2024nougat} to initialize the Swin Transformer of alignment encoder and image encoder.
The layer numbers and window size are {2, 2, 14, 2} and 7.
The hidden size of each layer is 1024 and the patch size is 4.
The input image size is 896 × 672.
We follow the vanilla Transformer-base \citep{vaswani2017attention} setting and pre-train an English-Chinese translation model on the UN Corpus.
We set the decoder’s max length and max position embeddings to 1536 to cover most input texts.
For the MLLM, we select four MLLMs with different numbers of parameters and training data: Vary-toy \citep{wei2024small}, Vary-base \citep{wei2023vary}, Llava-next \citep{liu2024llavanext} and Textmonkey \citep{liu2024textmonkey}, as shown in Table~\ref{table: mllms}.
The hyperparameter $\alpha$ is set to 1.0 and the sequence lengths for all MLLMs, except Llava-next, are set to 2048 to cover the long context of document images.
The sequence length for Llava-next is 4096 due to the different image encoders and prompts used by the MLLMs.

\begin{table}[t]
\footnotesize
\centering
\begin{tabular}{c|l|c|c}
\toprule
\multicolumn{1}{l|}{} & \textbf{MLLM}       & \textbf{\# Params (M)} & \textbf{Training Data} \\ \midrule
1                     & Vary-toy   & 2237.4        & Document images  \\
2                     & Vary-base  & 8123.7        & Document images  \\
3                     & Llava-next & 8354.8        & Natural images   \\
4                     & Textmonkey & 9715.8        & Document images  \\ \bottomrule
\end{tabular}
\caption{The number of parameters and training data of different MLLMs.}
\label{table: mllms}
\end{table}

During translation model pre-training, the maximum training step is 100K and the maximum token per batch is 4096.
A linear decay learning rate schedule with a learning rate of 7e-4 and a warmup ratio of 0.05 is used.
During the training stage of M4Doc, the maximum number of training steps is 15K and the batch size is 64.
We use a linear decay learning rate schedule with a learning rate of 5e-5 and the number of warmup steps is 1000.
We use Adam optimizer with $\beta_1 = 0.9$, $\beta_2 = 0.999$, $\epsilon = 1e-8$ for both training stages.
We used two NVIDIA A100 GPUs and spent 28 hours to complete the training of M4Doc (Vary-toy), and 68 hours to complete the training of M4Doc (Textmonkey).
For inference, we use beam search with 4 beams.

\begin{table}[t]
\footnotesize
\centering
\begin{tabular}{ll|ccc}
\toprule
                       &                        & \textbf{BLEU}  & \textbf{BLEU-PT} & \textbf{STEDS} \\ \midrule
\multicolumn{1}{l|}{1} & Cross-entropy   & 31.67 & 34.60   & 81.60 \\
\multicolumn{1}{l|}{2} & MSE               & 33.00 & 35.67   & 82.68 \\
\multicolumn{1}{l|}{3} & Cosine-similarity & 40.05 & 42.58   & 83.93 \\ \bottomrule
\end{tabular}
\caption{Results on DoTA English-Chinese valid set with different alignment loss functions.}
\label{table: loss function}
\end{table}

\begin{table*}[t]
\footnotesize
\centering
\begin{tabular}{ll|ccc|ccc|cc}
\toprule
\multicolumn{2}{c|}{\multirow{2}{*}{}}    & \multicolumn{3}{c|}{\textbf{DoTA}} & \multicolumn{3}{c|}{\textbf{DITrans}} & \multirow{2}{*}{\begin{tabular}[c]{@{}c@{}}\textbf{\# Params}\\ \textbf{(M)}\end{tabular}} & \multirow{2}{*}{\begin{tabular}[c]{@{}c@{}}\textbf{Time}\\ \textbf{(s/page)}\end{tabular}} \\
\multicolumn{2}{c|}{}                     & \textbf{BLEU}   & \textbf{BLEU-PT}  & \textbf{STEDS} & \textbf{BLEU}          & \textbf{BLEU-PT}         & \textbf{STEDS}         &                                                                         &                                                                          \\ \midrule
\multicolumn{1}{l|}{1} & Vary-toy (Original)         & 10.64  & 4.92    & 66.23 & 2.07          & 2.10            & 45.12         & 2237.4                                                                  & 62.58                                                                     \\
\multicolumn{1}{l|}{2} & Vary-toy (Fine-tuned)         & 22.67  & 22.75    & 73.99 & 5.87          & 6.14            & 49.59         & 2253.9                                                                  & 57.60                                                                     \\
\multicolumn{1}{l|}{3} & M4Doc (Vary-toy) & 39.95  & 42.33    & 83.97 & 14.79         & 18.67           & 53.73         & 212.4                                                                   & 9.61                                                                     \\ \midrule
\multicolumn{1}{l|}{4} & Vary-base (Original)             & 13.45  & 5.79    & 76.26 & 2.84         & 2.79           & 56.21         & 8123.7                                                                  & 68.84                                                                    \\
\multicolumn{1}{l|}{5} & Vary-base (Fine-tuned)             & 38.60  & 38.53    & 82.95 & 11.61         & 11.72           & 54.59         & 8137.9                                                                  & 69.67                                                                    \\
\multicolumn{1}{l|}{6} & M4Doc (Vary-base)     & 41.22  & 42.09    & 86.06 & 14.52         & 16.55           & 55.89         & 215.6                                                                   & 9.43                                                                     \\ \bottomrule
\end{tabular}
\caption{Results of directly fine-tuning MLLMs on the DoTA dataset. \textbf{\# Params} is the number of parameters of the model during inference. \textbf{Time} is the average inference time on a single NVIDIA V100 GPU.}
\label{table: fine-tune mllm}
\end{table*}

\subsection{Prompts for Each MLLM}
The \texttt{<System Prompt>} and \texttt{<User Prompt>} used in the main experiment are listed as follows.

\begin{tcolorbox}[colback=lightgray!50!white,colframe=lightgray,title=\textcolor{black}{Prompts for Vary-toy/base}, width=\columnwidth, breakable]
\footnotesize
\texttt{<System Prompt>} \newline
None \newline \newline
\texttt{<User Prompt>} \newline
Convert the image to markdown/latex format.
\end{tcolorbox}

\begin{tcolorbox}[colback=lightgray!50!white,colframe=lightgray,title=\textcolor{black}{Prompts for Llava-next}, width=\columnwidth, breakable]
\footnotesize
\texttt{<System Prompt>} \newline
You are a helpful language and vision assistant. You are able to understand the visual content that the user provides, and assist the user with a variety of tasks using natural language. \newline \newline
\texttt{<User Prompt>} \newline
OCR this image.
\end{tcolorbox}

\begin{tcolorbox}[colback=lightgray!50!white,colframe=lightgray,title=\textcolor{black}{Prompts for Textmonkey}, width=\columnwidth, breakable]
\footnotesize
\texttt{<System Prompt>} \newline
You are a helpful assistant. \newline \newline
\texttt{<User Prompt>} \newline
Read all the text in the image.
\end{tcolorbox}

\section{Detailed Analysis}
\subsection{Effect of Different Alignment Loss Functions}
\label{sec: loss function}
To explore the impact of different alignment loss functions, we use cross-entropy loss, mean square error (MSE) loss, and cosine-similarity loss as the alignment loss function and conduct experiments with the same setting as the main experiment M4Doc (Vary-toy).
The results are shown in Table~\ref{table: loss function}.

As shown in the table, using cosine similarity as the alignment loss function yields the best results.
We think this may be because the loss values calculated by cosine-similarity range between $[-1, 1]$, allowing the model to strike a balance between learning the alignment task and the translation task.
Therefore, we choose cosine-similarity loss for the main experiment.

\subsection{Hyperparameter Sensitivity Analysis}
\label{sec: hyperparameter}
\begin{figure}[t]
    \centering
    \includegraphics[width=\columnwidth]{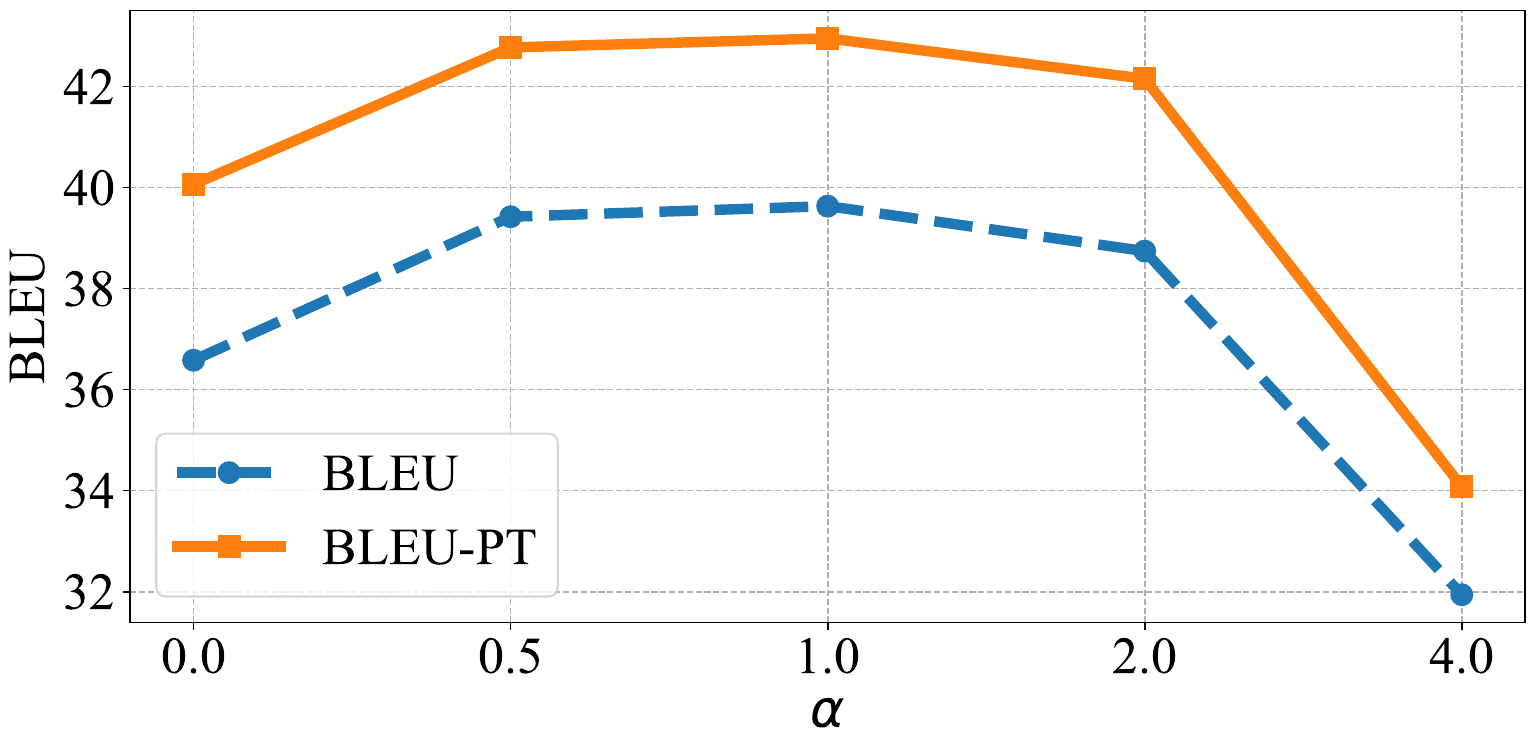}
    \caption{BLEU and BLEU-PT scores of M4Doc (Vary-toy) trained with different $\alpha$ values on the valid set. Detailed data can be seen in Appendix~\ref{sec: appendix detailed data}.}
    \label{figure: alpha}
\end{figure}

To explore the impact of $\alpha$ in the loss function, we vary $\alpha$ and get results in Figure~\ref{figure: alpha}.
As shown in the figure, the model's performance initially increases and then decreases with the increase in $\alpha$, achieving the best performance when $\alpha = 1.0$.
This could potentially be attributed to the fact that a small $\alpha$ diminishes the influence of MLLM, while a large $\alpha$ introduces too much noise.
So, we set $\alpha = 1.0$ in the main experiment.

\subsection{Comparison with Fine-tuning MLLM}
\label{sec: appendix fine-tune mllm}
We conduct comparative experiments to evaluate the DIMT capabilities of MLLMs by directly applying MLLMs to the DIMT task and fine-tuning them specifically for this task.  
We fine-tune Vary-toy and Vary-base using LoRA \citep{DBLP:conf/iclr/HuSWALWWC22} with a $lora\_rank$ of 32, while keeping other settings consistent with the main experiment.  
The results are presented in Table~\ref{table: fine-tune mllm}.

As shown in the table, directly using MLLMs for the DIMT task yields very poor performance (line 1 and 4), with almost no translation capability on the political report domain.
After fine-tuning on the DoTA dataset, the DIMT capability of MLLMs improves significantly but still falls short of the performance achieved by our proposed M4Doc method.
By comparing line 5 and 6, our method outperforms the best-performing Vary-base (Fine-tuned) model by 2.62 BLEU scores and achieves a greater improvement of 2.91 BLEU scores in zero-shot cross-domain scenarios.
This highlights the potential of our method for efficiently leveraging MLLMs across various downstream tasks.

\begin{table}[t]
\footnotesize
\centering
\resizebox{\linewidth}{!}{
\begin{tabular}{cl|cccc}
\toprule
\multicolumn{2}{c|}{}                       & \textbf{B}  & \textbf{C} & \textbf{BP} & \textbf{S} \\ \midrule
\multicolumn{1}{c|}{1} & GPT-4o             & 29.17 & 60.32 & 31.95   & 59.45 \\
\multicolumn{1}{c|}{2} & Gemini             & 30.31 & 59.67 & 31.69   & 63.32 \\ \midrule
\multicolumn{1}{c|}{3} & DIMTDA             & 38.73 & 61.33 & 42.37   & 84.98 \\
\multicolumn{1}{c|}{4} & M4Doc (Vary-toy)   & 39.45 & 62.42 & 42.59   & 83.50 \\
\multicolumn{1}{c|}{5} & M4Doc (Vary-base)  & 41.11 & 63.57 & 42.00   & 86.62 \\
\multicolumn{1}{c|}{6} & M4Doc (Textmonkey) & 42.27 & 65.40 & 44.17   & 86.93 \\ \bottomrule
\end{tabular}
}
\caption{Results on comparison with commercial MLLMs. \textbf{B}, \textbf{C}, \textbf{BP}, and \textbf{S} represent BLEU, COMET, BLEU-PT, and STEDS, respectively.}
\label{table: commercial mllms}
\end{table}

\subsection{Comparison with Commercial MLLMs}
With the rapid development of MLLMs, some commercial MLLMs \citep{hurst2024gpt, team2024gemini} demonstrate the capability of understanding text-rich document images.
To assess their ability to accomplish the DIMT task, we randomly choose 200 samples from the test set of the DoTA dataset, then prompt GPT-4o and Gemini with three different prompts to complete the document image machine translation task.
The prompts we used are as follows.

\begin{tcolorbox}[colback=lightgray!50!white,colframe=lightgray,title=\textcolor{black}{Prompts for GPT-4o and Gemini to complete DIMT task}, width=\columnwidth, breakable]
\footnotesize
\texttt{<Prompt 1>} \newline
Output the Chinese translations of this image in markdown format. \newline \newline
\texttt{<Prompt 2>} \newline
Please extract and provide the Chinese translations of the text contained within this image, ensuring that the translations are accurately represented, and format them using markdown for clear presentation. \newline \newline
\texttt{<Prompt 3>} \newline
Please translate the all texts in this image into English and adhere to the following translation standards: \newline
Accuracy: Ensure that the translation fully captures the meaning of all the texts in the image without adding or omitting any information. \newline
Fluency: The translation should read naturally and smoothly, reflecting the conventions of the target language and the translation should follow the reading order of the image. \newline
Format: The translation should be presented in markdown format.
\end{tcolorbox}

We average the metric values of the translation results obtained from different prompts to determine the final results.
As the output format of MLLMs may be unstable, we filter the English parts of the output text and only keep the Chinese parts.

Table~\ref{table: commercial mllms} reveals that both GPT-4o and Gemini can accomplish the DIMT task directly, but exhibit inferior performance compared to M4Doc (line 2 vs. line 6).
This may be because commercial MLLMs are not trained on the DoTA dataset, their output formats differ from the reference.
This leads to commercial MLLMs performing significantly worse than M4Doc on metrics like BLEU and STEDS.
However, semantic-based evaluation metrics, such as COMET, can more accurately reflect the model's translation performance, which shows that the DIMT ability of existing commercial MLLMs is comparable to that of M4Doc.

\subsection{Evaluation on Other Languages}
To verify our method's effectiveness in other languages, we conduct English-French and English-German DIMT experiments.
The text machine translation models are pre-trained on the UN Corpus En-Fr and WMT14 En-De dataset.
We use the En-Fr and En-De subsets of the DoTA dataset to train our models.
The rest of the settings remain the same as the main experiment.
Table~\ref{table: multilingual} demonstrates the effectiveness of M4Doc on other languages' DIMT tasks.

\begin{table}[t]
\footnotesize
\centering
\resizebox{\linewidth}{!}{
\begin{tabular}{llcccc}
\toprule
\multicolumn{2}{c|}{\multirow{2}{*}{}}                           & \multicolumn{2}{c|}{\textbf{En-Fr}}         & \multicolumn{2}{c}{\textbf{En-De}} \\
\multicolumn{2}{c|}{}                                            & \textbf{BLEU}  & \multicolumn{1}{c|}{\textbf{STEDS}} & \textbf{BLEU}        & \textbf{STEDS}       \\ \midrule
\multicolumn{1}{l|}{1} & \multicolumn{1}{l|}{Text-only MT}       & 59.68 & \multicolumn{1}{c|}{95.93} & 49.25       & 96.04       \\ \midrule
\multicolumn{6}{c}{\textbf{Cascade Baselines}} \\ \midrule
\multicolumn{1}{l|}{2} & \multicolumn{1}{l|}{LARDIT}             & 42.79 & \multicolumn{1}{c|}{75.59} & 32.65       & 75.59       \\
\multicolumn{1}{l|}{3} & \multicolumn{1}{l|}{Nougat-trans}       & 55.82 & \multicolumn{1}{c|}{90.77} & 43.73       & 89.92       \\ \midrule
\multicolumn{6}{c}{\textbf{End-to-end DIMT}} \\ \midrule
\multicolumn{1}{l|}{4} & \multicolumn{1}{l|}{DIMTDA}             & 45.82 & \multicolumn{1}{c|}{84.84} & 37.83       & 85.92       \\
\multicolumn{1}{l|}{5} & \multicolumn{1}{l|}{M4Doc (Vary-toy)}   & 48.88 & \multicolumn{1}{c|}{85.04} & 41.47       & 86.72       \\
\multicolumn{1}{l|}{6} & \multicolumn{1}{l|}{M4Doc (Vary-base)}  & 49.18 & \multicolumn{1}{c|}{86.83} & 42.61       & 86.64       \\
\multicolumn{1}{l|}{7} & \multicolumn{1}{l|}{M4Doc (Textmonkey)} & 54.64 & \multicolumn{1}{c|}{89.85} & 46.70       & 89.58       \\ \bottomrule
\end{tabular}
}
\caption{Results on DoTA English-French and English-German test set.}
\label{table: multilingual}

\end{table}

\section{Detailed Data}
\label{sec: appendix detailed data}
Table~\ref{table: context length} presents the detailed data corresponding to the BLEU scores of M4Doc models tested on validation sets with different context lengths, as shown in Figure~\ref{figure: context length}.  
Table~\ref{table: alpha} provides the detailed data corresponding to the BLEU and BLEU-PT scores of M4Doc (Vary-toy) trained with different $\alpha$ values on the validation set, as illustrated in Figure~\ref{figure: alpha}.

\begin{table}[h]
\footnotesize
\centering
\begin{tabular}{l|cc}
\toprule
                   & \textbf{(0,250]}   & \textbf{(250,500]} \\ \midrule
DIMTDA             & 51.13              & 45.95              \\
M4Doc (Vary-toy)   & 51.82              & 45.75              \\
M4Doc (Vary-base)  & 52.85              & 49.46              \\
M4Doc (Textmonkey) & 56.04              & 50.15              \\ \midrule
                   & \textbf{(500,750]} & \textbf{(750,)}    \\ \midrule
DIMTDA             & 39.98              & 34.85              \\
M4Doc (Vary-toy)   & 45.54              & 41.10              \\
M4Doc (Vary-base)  & 45.43              & 39.02              \\
M4Doc (Textmonkey) & 48.65              & 45.63              \\ \bottomrule
\end{tabular}
\caption{Detailed data of Figure~\ref{figure: context length}.}
\label{table: context length}

\end{table}

\begin{table}[h]
\footnotesize
\centering
\begin{tabular}{c|cc}
\toprule
$\alpha$ & \textbf{BLEU}  & \textbf{BLEU-PT} \\ \midrule
0.0   & 36.58 & 40.06   \\
0.5   & 39.42 & 42.77   \\
1.0   & 39.63 & 42.95   \\
2.0   & 38.74 & 42.15   \\
4.0   & 31.94 & 34.08   \\ \bottomrule
\end{tabular}
\caption{Detailed data of Figure~\ref{figure: alpha}.}
\label{table: alpha}

\end{table}

\section{Output Samples}
\label{sec: appendix output samples}
\begin{figure*}[h]
    \centering
    \includegraphics[width=2\columnwidth]{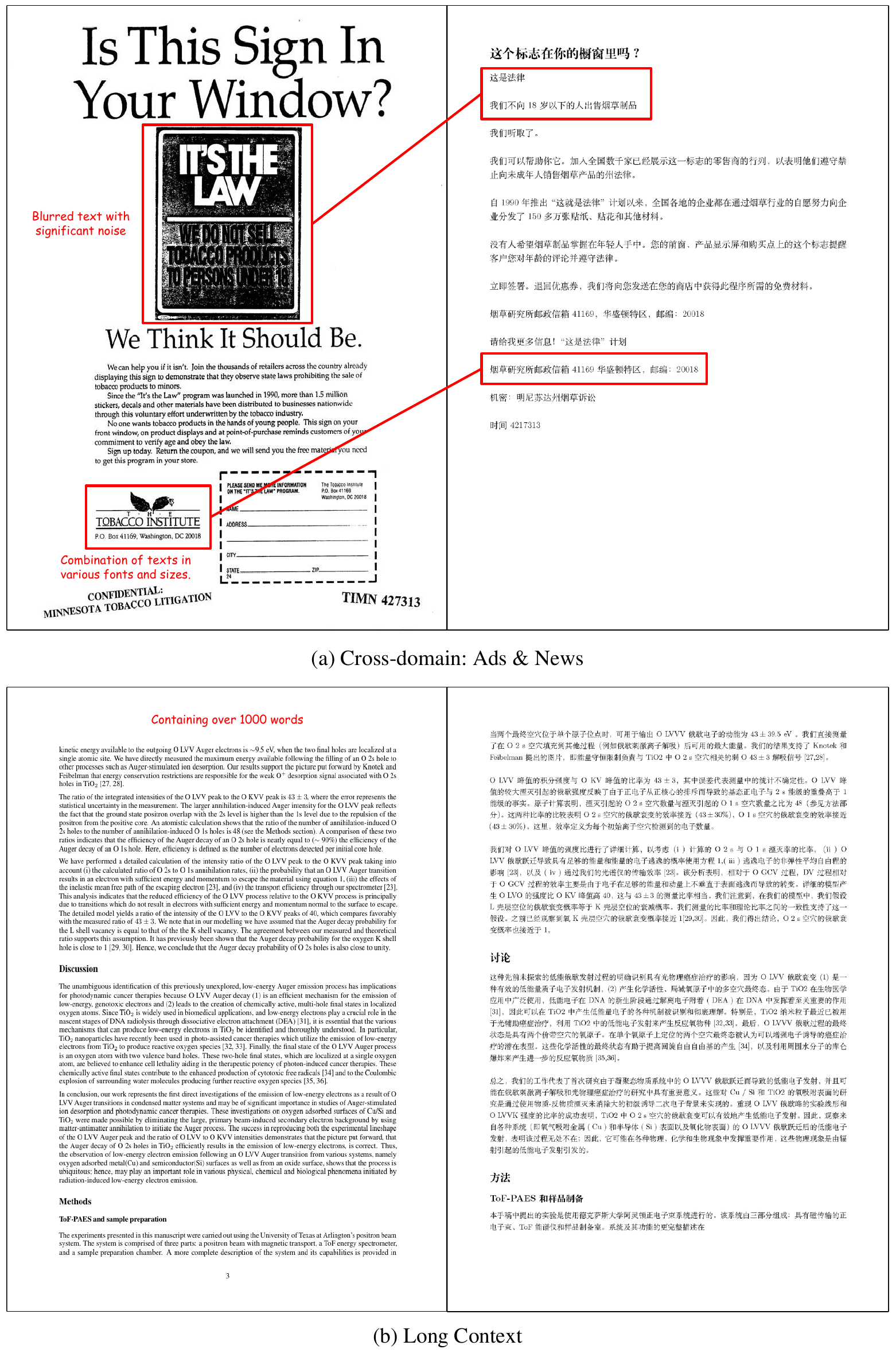}
    \caption{The output samples of M4Doc. For each image pair, the left is the input document image, and the right is the output translations in markdown format after rendering. It is better to zoom in for a clearer view.}
    \label{figure: appendix case}
\end{figure*}

We provide the output samples of M4Doc in cross-domain and long context scenarios in Figure~\ref{figure: appendix case}.
Figure~\ref{figure: appendix case} (a) is an image from the ads \& news subset of the DITrans dataset.
The scanned document image contains a lot of noise, and the font size varies significantly, which makes the image difficult to handle.
After fine-tuning our model on the subset, it can translate the text in the image, even if some of the text appears blurry.

Figure~\ref{figure: appendix case} (b) is an image from the DoTA dataset, which contains more than 1000 English words.
For images containing such long contexts, our model still achieves end-to-end DIMT without omissions.

\begin{figure*}[h]
    \centering
    \includegraphics[width=2\columnwidth]{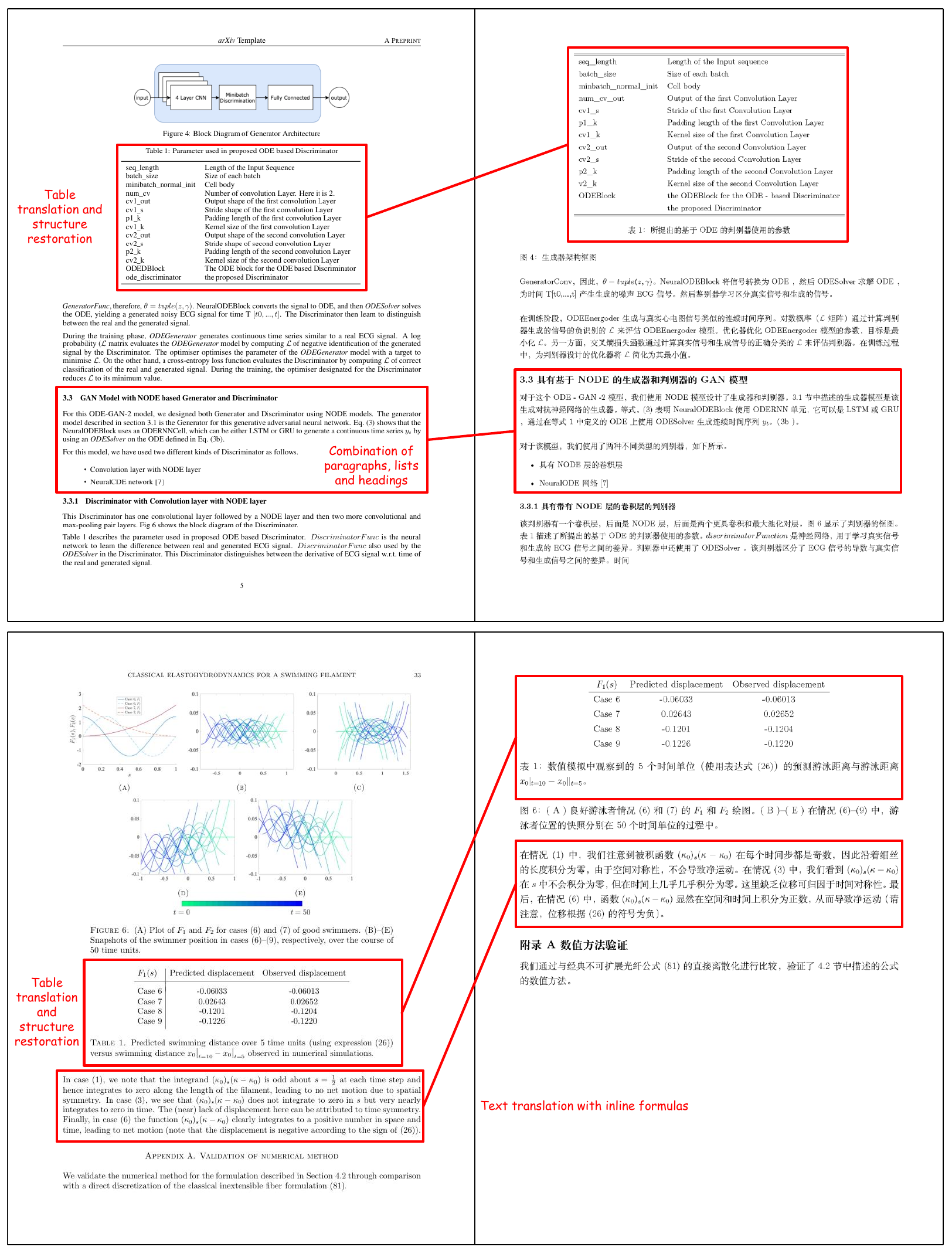}
    \caption{The output samples of M4Doc in the DoTA dataset. It is better to zoom in for a clearer view.}
    \label{figure: appendix dota}
\end{figure*}

\begin{figure*}[h]
    \centering
    \includegraphics[width=2\columnwidth]{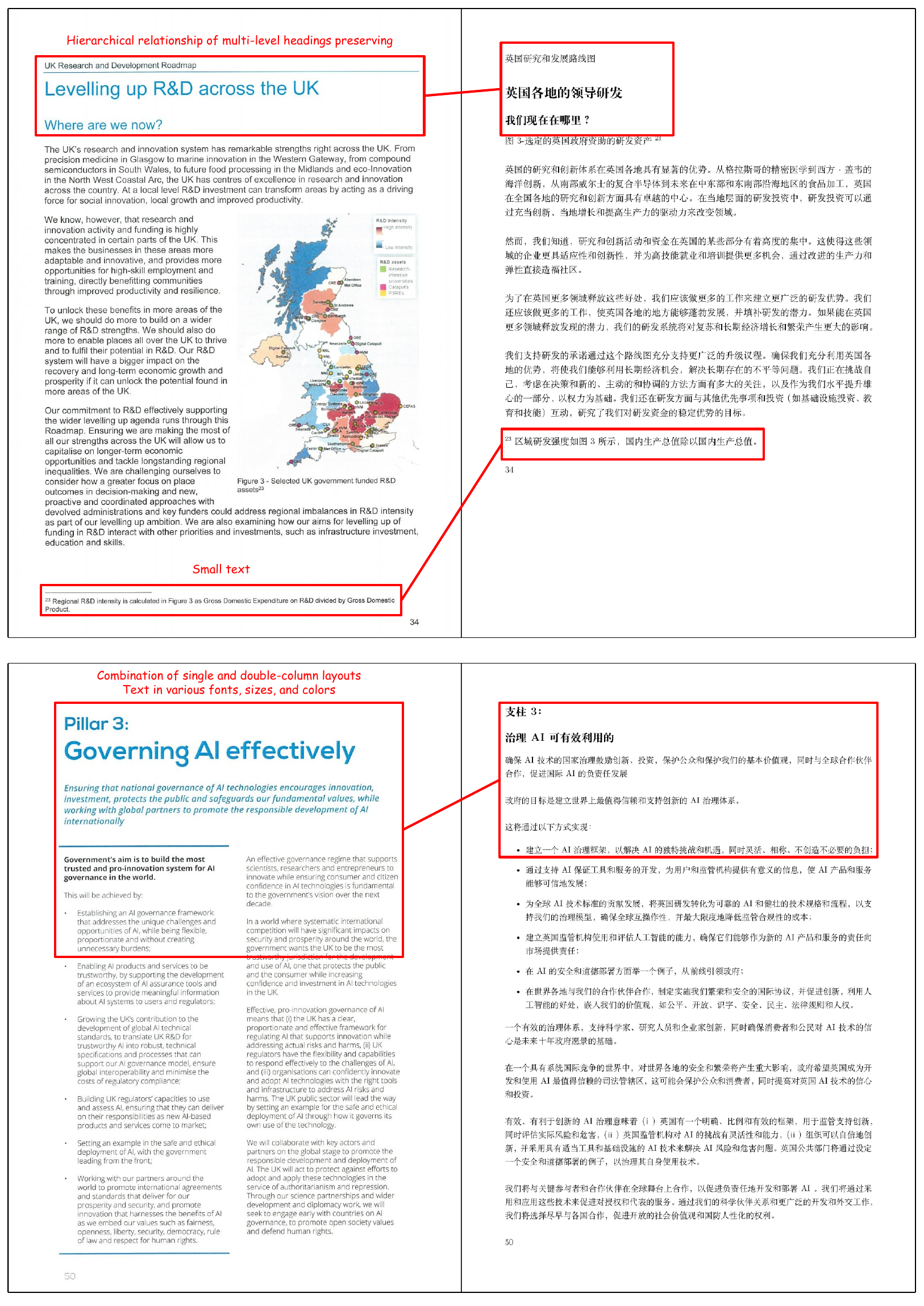}
    \caption{The output samples of M4Doc in the political report subset of the DITrans dataset. It is better to zoom in for a clearer view.}
    \label{figure: appendix political report}
\end{figure*}

\begin{figure*}[h]
    \centering
    \includegraphics[width=2\columnwidth]{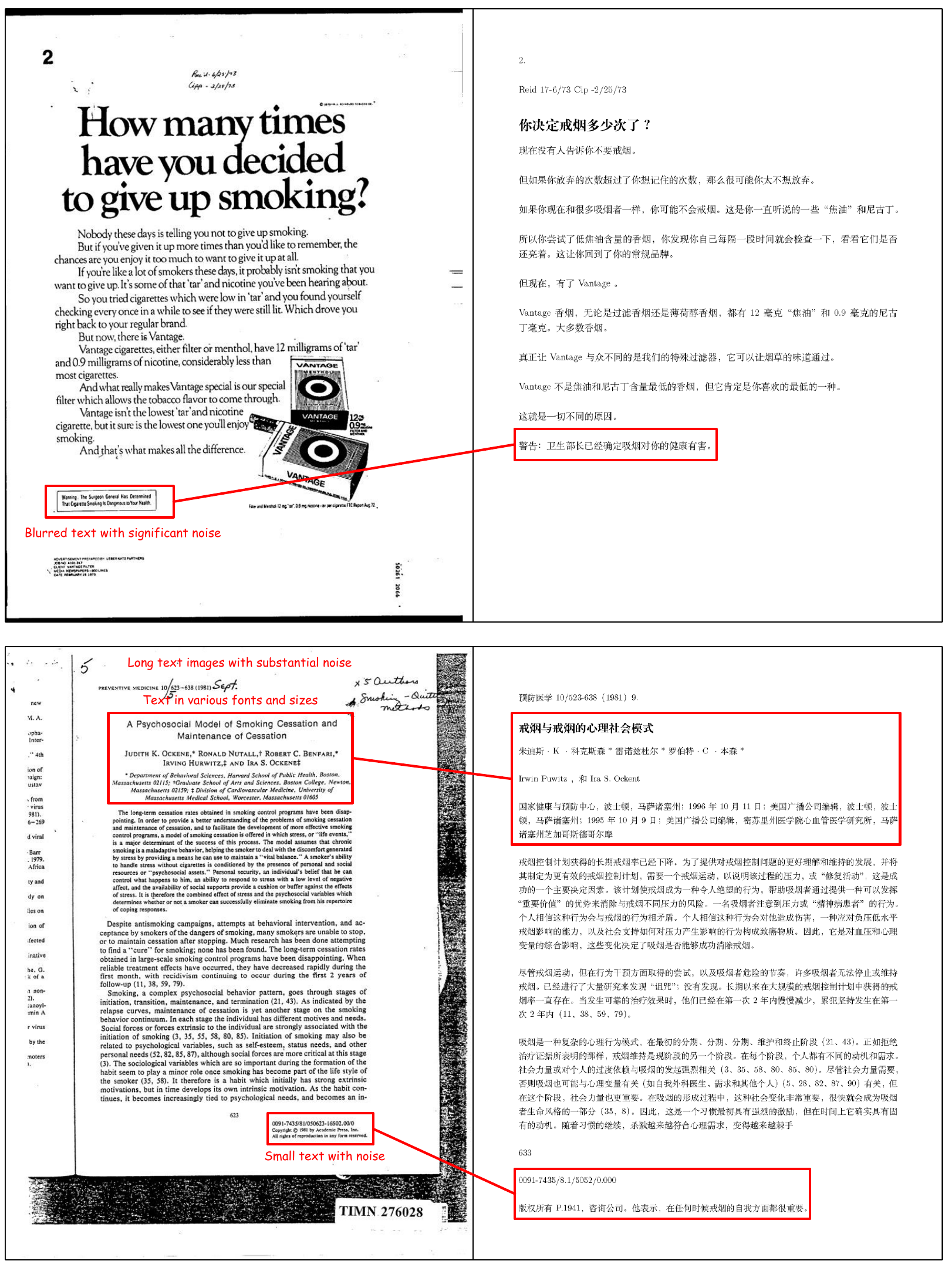}
    \caption{The output samples of M4Doc in the ads \& news subset of the DITrans dataset. It is better to zoom in for a clearer view.}
    \label{figure: appendix ads news}
\end{figure*}

We also list other output samples in Figure~\ref{figure: appendix dota}, Figure~\ref{figure: appendix political report}, and Figure~\ref{figure: appendix ads news}.

\end{document}